\documentclass{article}

\usepackage{PRIMEarxiv}

\usepackage[utf8]{inputenc} % allow utf-8 input
\usepackage[T1]{fontenc}    % use 8-bit T1 fonts
\usepackage{amsmath,amsfonts,amssymb,bm}
\usepackage[numbers,sort&compress]{natbib}
\usepackage{hyperref}       % hyperlinks
\usepackage{url}            % simple URL typesetting
\usepackage{booktabs}       % professional-quality tables
\usepackage{nicefrac}       % compact symbols for 1/2, etc.
\usepackage{microtype}      % microtypography
\usepackage{graphicx}       % graphics
\usepackage{array}
\usepackage{tabularx}
\usepackage{float}
\usepackage{algorithm}
\usepackage{algpseudocode}
\usepackage[skins,breakable]{tcolorbox}

\makeatletter
\renewenvironment{figure*}[1][tbp]{%
  \@float{figure}[#1]%
}{%
  \end@float
}
\makeatother

\usepackage{amsmath,amsfonts,bm}

\def\eqref#1{equation~\ref{#1}}
\def\Eqref#1{Equation~\ref{#1}}
\def\1{\bm{1}}

\DeclareMathAlphabet{\mathsfit}{\encodingdefault}{\sfdefault}{m}{sl}
\SetMathAlphabet{\mathsfit}{bold}{\encodingdefault}{\sfdefault}{bx}{n}

\tcbset{
  promptbox/.style={
    enhanced,
    breakable,
    colback=black!4,
    colframe=black!45,
    colbacktitle=black!8,
    coltitle=black,
    fonttitle=\bfseries,
    boxrule=0.4pt,
    arc=2pt,
    left=8pt,
    right=8pt,
    top=5pt,
    bottom=5pt,
    before skip=8pt,
    after skip=10pt,
  }
}

\title{EvoDisco: Strategy Discovery for Evolving Agent Memory
\thanks{\textit{\underline{Citation}}:
\textbf{Liu et al. EvoDisco: Strategy Discovery for Evolving Agent Memory. Preprint.}}
}

\author{%
  \normalfont
  \textit{Mingyang} \textsc{Liu}$^{1}$,
  \textit{Qingcan} \textsc{Kang}$^{2}$,
  \textit{Shixiong} \textsc{Kai}$^{2}$,
  \textit{Kaichao} \textsc{Liang}$^{2}$\\
  \normalfont
  \textit{Xiaojin} \textsc{Fu}$^{2}$,
  \textit{Tao} \textsc{Zhong}$^{2}$,
  \textit{Linqi} \textsc{Song}$^{1}$\\[4pt]
  \normalfont
  $^{1}$City University of Hong Kong\\
  \normalfont
  $^{2}$Huawei Technologies Ltd.%
}

\begin{document}
\maketitle

\begin{abstract}
Large language model agents accumulate experience in memory, and how that memory evolves during interaction is as important as what it stores. Recent methods update memory online, yet both the operators that transform memory and the rules that select them remain fixed by design. We propose \textbf{EvoDisco}, which formulates agent memory evolution as \emph{evolutionary strategy discovery}: the agent must discover which operator to apply, when to apply it, and how to instantiate it, rather than rewriting memory under a static operator schedule. A learnable policy conditions on the current memory state and recent task history, then selects and instantiates one of four semantic operators---\textsc{Abstract}, \textsc{Refine}, \textsc{Bridge}, and \textsc{Reframe}---which respectively compress related memories, decompose overloaded entries, connect weakly linked experiences, and reorganize retrieved reasoning chains. The policy is updated with a bi-level fitness. An inner signal scores each transformation using a retrieval-quality improvement regularized by memory complexity. An outer signal scores strategic efficacy over a window of past decisions, combining historical operator-utility regret, utility-prediction error, and selection diversity. Under a one-pass online plus frozen-transfer protocol on ALFWorld, WebShop, and LifelongAgentBench (OS and DB), EvoDisco improves over strong memory baselines including MemP and MemRL. The largest gains are on ALFWorld ($+8.8$ online / $+12.9$ frozen-transfer points over MemP; $+16.4$ / $+17.9$ over MemRL). Ablations attribute these gains to both fitness levels and the operator library.
\end{abstract}


% ============================================================
% 1. INTRODUCTION
% ============================================================
\section{Introduction}
\label{sec:introduction}

Large Language Model (LLM)-based agents have emerged as a powerful paradigm for tackling complex, long-horizon tasks that require reasoning, planning, and multi-turn interaction. However, effective agency in such settings demands more than the ability to solve isolated prompts---agents must accumulate experience, retain task-relevant information across sessions, and selectively reuse past knowledge. This requirement exposes a fundamental limitation of context-only interaction: even with long-context models, relevant information can be diluted or forgotten as interactions grow, leading to unstable recall and degraded reasoning performance.

Memory systems have thus become a core module in LLM-based agents. Recent work has increasingly recognized that \emph{how} memory is organized matters as much as \emph{what} is stored. A growing body of research has moved beyond flat vector retrieval toward structured memory representations, particularly knowledge graphs, which offer superior capabilities for modeling inter-memory relations, enabling multi-hop reasoning, and providing interpretable access. Representative systems include AriGraph~\citep{arigraph}, which integrates semantic and episodic memories within a unified knowledge graph; Zep~\citep{zep}, which provides a temporal knowledge graph architecture for agent memory; A-MEM~\citep{amem}, which adopts a Zettelkasten-inspired interconnected knowledge network; and G-Memory~\citep{gmemory}, which employs a hierarchical graph-based memory design. These graph-based approaches have demonstrated significant advantages over unstructured memory in tasks requiring structured access and associative retrieval.

Despite these advances, a critical limitation persists across existing memory systems: \textbf{the memory structure itself remains static after its initial construction}. Most structured memory mechanisms rely on one-time restructuring workflows to build their memories, neglecting ongoing updates and optimization of the memory structures. An advanced agent should progressively enrich and refine its memory structure throughout interactions with the environment. While some recent works have begun to address memory evolution---for instance, HAGE~\citep{jiang2026hage} employs reinforcement learning to drive weighted graph evolution, and MemEvolve~\citep{zhang2026memevolve} introduces a meta-evolutionary framework that jointly evolves experiential knowledge and memory architecture---these approaches still rely on \emph{fixed evolutionary operators} (e.g., mutation and crossover with predetermined probabilities) or \emph{manually engineered adaptation mechanisms}.

This raises a fundamental question: \textbf{If the agent itself is capable of reasoning about its own memory, why should the \emph{way} memory evolves be predetermined by human designers?}

In this paper, we propose \textbf{EvoDisco} (Strategy Discovery for Evolving Agent Memory), a novel framework that elevates agent memory evolution from \emph{rewriting memory under a fixed operator schedule} to \emph{discovering how memory should evolve}. Unlike prior work that treats evolution as a blind search using fixed genetic operators, EvoDisco enables the agent to \emph{discover which evolutionary operation to apply, when to apply it, and how to instantiate it} based on the current memory state and recent task performance. The agent maintains a \textbf{strategy policy}---a learnable module that selects from a curated cognitive operator library (abstraction, refinement, bridging, and reframing)---and the \emph{policy itself} undergoes evolution guided by a bi-level fitness function that evaluates both the immediate utility of each operation and the strategic efficacy of the agent's historical decisions.
Figure~\ref{fig:framework} gives an overview.

Our contributions are threefold:

\begin{itemize}
    \item \textbf{Evolutionary Strategy Discovery.} We formulate memory evolution as a learnable strategy-selection problem, shifting from a fixed operator schedule to state-conditioned selection and instantiation of semantic memory transformations.

    \item \textbf{Bi-Level Fitness with Strategic Efficacy.} We introduce a dual fitness framework that evaluates the retrieval-quality change and structural cost of each memory update, together with the policy's historical accuracy in selecting high-utility operators.

    \item \textbf{Cognitive Operator Library.} We design a curated set of four semantic-level operators---\textsc{Abstract} (compressing related memories into higher-level concepts), \textsc{Refine} (decomposing overloaded memories), \textsc{Bridge} (connecting weakly linked experiences), and \textsc{Reframe} (reorganizing retrieved reasoning chains)---that provide sufficient expressive power while maintaining controllability.
\end{itemize}

Through extensive experiments on ALFWorld, WebShop, and LifelongAgentBench (OS and DB), we evaluate the learned policy against fixed-strategy and memory baselines.

% ============================================================
% 2. RELATED WORK
% ============================================================
\section{Related Work}
\label{sec:related_work}

\subsection{Memory Systems for LLM Agents}
\label{sec:rel_memory_systems}

Memory lets an agent retain experience beyond the current context. Early systems store transcripts, embeddings, or key-value records and retrieve them by semantic similarity~\citep{lewis2020retrieval}. Similarity search is effective for direct lookup, but it provides little support for multi-hop relations or for deciding how stored experience should be reorganized.

Graph-structured memory addresses the relational part of this problem. GraphRAG-style pipelines extract entities and relations and use the resulting graph for retrieval and summarization~\citep{graphrag,sgprompt,holmes,graphreader}. Agent memory systems then keep such graphs online: AriGraph~\citep{arigraph} links semantic and episodic memories, Zep~\citep{zep} indexes a temporal knowledge graph, A-MEM~\citep{amem} grows a Zettelkasten-style note network, and G-Memory~\citep{gmemory} organizes memory hierarchically for multi-agent settings. Other systems do not rely on an explicit graph. Mem0 merges atomic facts~\citep{chhikara2025mem0}, MemP distills trajectories into procedures~\citep{fang2025memp}, Self-RAG gates retrieved text~\citep{asai2024selfrag}, and MemRL ranks episodic notes with a nonparametric $Q$-value while the backbone stays frozen~\citep{zhang2026memrl}. Surveys organize these designs by structure and function~\citep{survey_memory}. A separate limitation cuts across them: most structured memories are built by a one-time workflow and are not subsequently reorganized as the task distribution changes~\citep{survey_memory_evolution}.

\begin{figure*}[t]
    \centering
    \includegraphics[width=\textwidth]{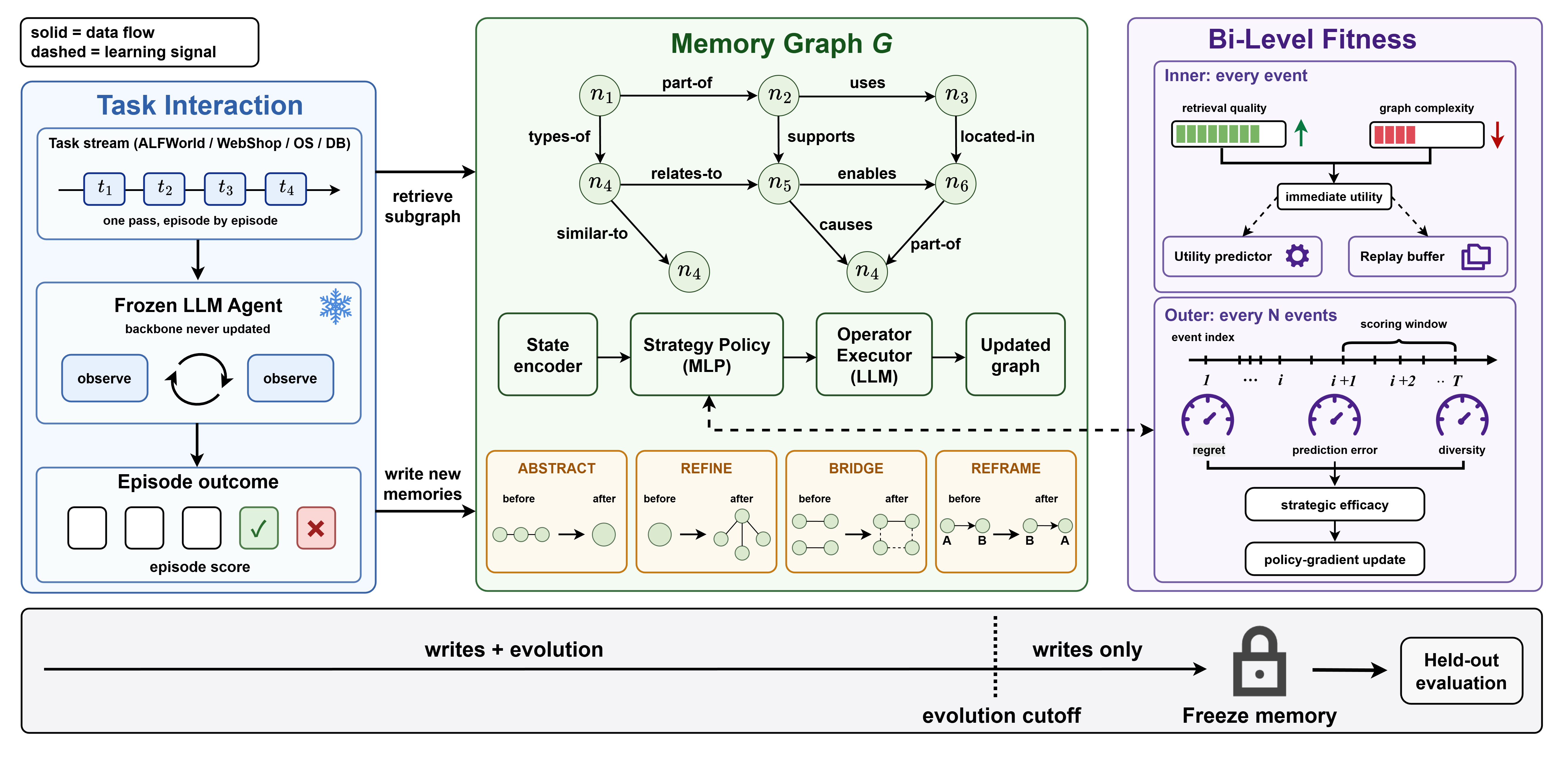}
    \caption{
        Overview of EvoDisco.
        A frozen language model interacts with a one-pass task stream,
        retrieves a subgraph, and writes new memories after each episode.
        A strategy policy reads the memory-graph state and selects one of
        four operators---\textsc{Abstract}, \textsc{Refine}, \textsc{Bridge},
        and \textsc{Reframe}---which the operator executor applies to the graph.
        Immediate utility scores each edit by the change in retrieval quality
        regularized by graph complexity.
        Every $N$ events, strategic efficacy combines operator-utility regret,
        utility-prediction error, and selection diversity, and updates the policy.
        Evolution runs only before the cutoff; afterward the agent continues
        to write memories.
        Held-out evaluation freezes the accumulated memory.
    }
    \label{fig:framework}
\end{figure*}

\subsection{Memory Evolution}
\label{sec:rel_memory_evolution}

Several systems do update memory over time. Production agents already separate declarative, procedural, and episodic stores and run background consolidation to distill or prune them~\citep{memgpt}. In the research literature, MemEvolve jointly evolves experiential knowledge and the memory architecture~\citep{zhang2026memevolve}; EvolveMem treats retrieval configuration as an action space tuned by an LLM diagnosis module~\citep{evolvemem}; HAGE trains edge weights on a graph from downstream task signals~\citep{jiang2026hage}; MemRL keeps the topology fixed and updates only the utility scores used at retrieval~\citep{zhang2026memrl}; and FORGE evolves natural-language memory with a population-based protocol~\citep{Bogdanov_2026}. Surveys describe this line as a shift from storing trajectories, to reflecting on them, to abstracting them into reusable experience~\citep{survey_selfevolve,selfevolve_survey}.

What stays fixed is the evolutionary mechanism itself. Mutation and crossover use predetermined probabilities, edge updates follow a chosen reinforcement-learning rule, and configuration search stays inside a human-specified action space. The agent can change memory content, weights, or a small set of architecture parameters, but it does not learn which kind of reorganization is appropriate for the current memory state.

\subsection{Strategy Selection and Evolutionary Search}
\label{sec:rel_metacog}

Choosing among strategies is well studied when the strategy is an external behavior. SMART models strategy selection for reasoning as a Markov decision process~\citep{smart}. MetaClaw continually updates both a base policy and a library of reusable skills~\citep{metaclaw}. Meta-reinforcement learning gives a broader account of adapting from environmental feedback at test time~\citep{metarl}. These methods select an action in the task, not an edit to the agent's own knowledge structure. Metacognitive work on LLM agents similarly monitors answers and reasoning traces~\citep{metacog_llm}, rather than the rule used to reorganize memory.

Evolutionary computation supplies the other nearby toolset. Genetic programming and Genetic Network Programming evolve behaviors or graph-structured programs~\citep{genetic_prog,gnp}; LLM-in-the-loop methods mutate prompts or code~\citep{evoprompt,llm_gi}; and EvoLattice stores a population of candidate behaviors in one directed acyclic graph~\citep{evolattice}. The operators in these loops are part of the algorithm design. When evolution is applied to memory, it typically decides which experiences to retain, not how the memory graph should be restructured, and the operator set remains constant.

EvoDisco targets that missing choice. Relative to MemEvolve, the learned object is the policy that selects operators, not a schedule of architecture parameters. Relative to HAGE, the agent chooses among qualitatively different semantic edits rather than training one edge representation. Relative to SMART and MetaClaw, the action is an internal reorganization of memory. Bi-level fitness then scores both the immediate utility of an edit and the policy's historical accuracy in choosing high-utility operators.

% ============================================================
% 3. METHOD
% ============================================================
\section{Method}
\label{sec:method}

EvoDisco learns a state-dependent strategy for reorganizing
an agent's persistent memory.
Task interactions supply new memories, cognitive operators
transform their organization, and observed outcomes guide
subsequent operator selection.
The backbone language model remains frozen; adaptation occurs
through the memory graph and a lightweight strategy policy.
Figure~\ref{fig:framework} summarizes this loop.

\subsection{Problem Setup and Memory}
\label{sec:problem}
\label{sec:graph_rep}

An agent processes tasks drawn from a distribution $\mathcal{T}$
while maintaining a memory graph $\mathcal{G}=(\mathcal{V},\mathcal{E})$.
For a fixed graph, the population objective is
\begin{equation}
    F(\mathcal{G})
    =
    \mathbb{E}_{t\sim\mathcal{T}}
    \left[\mathrm{score}(t;\mathcal{G})\right].
    \label{eq:memory_objective}
\end{equation}
We distinguish this expectation from the observed score $y_j$
of task episode $j$.

Memory is a directed, multi-relational graph.
Nodes store facts, episode summaries, procedures, or concepts;
typed edges encode semantic and structural relations.
Each node $v$ stores an embedding $\mathbf{e}_v$,
a timestamp $t_v$, a confidence $c_v\in[0,1]$,
and a retrieval count $r_v$.
An edge is a typed relation $(u,\rho,v)$ with type
$\rho\in\mathcal{R}$, strength $w_e\in[0,1]$,
and evidence-recency $\tau_e$.
The fixed vocabulary includes
\texttt{causes}, \texttt{precedes}, \texttt{analogous-to},
\texttt{is-a}, and \texttt{generalizes}.
After each episode, an LLM-based extractor writes memory units
and candidate relations.
During task execution, query-centered spreading activation
retrieves a subgraph for the agent's context.

At evolution event $i$, the agent encodes the current graph
and recent history as a state $s_i$, then selects and executes
\begin{align}
    a_i &= (o_i,c_i,\theta_i),
    \qquad o_i\in\mathcal{O},
    \label{eq:evolution_action}\\
    \mathcal{G}_i^{+}
    &= \mathsf{T}_{o_i}(\mathcal{G}_i;c_i,\theta_i),
    \label{eq:graph_transition}
\end{align}
where $c_i$ indexes a graph candidate and $\theta_i$ is the instantiation.
Events index triggered attempts, including attempts that
leave the graph unchanged, rather than task episodes.
Ordinary memory writes may occur between events.
The state $s_i$ concatenates four groups:
graph statistics (node and edge counts, mean degree,
modularity, and path length);
retrieval statistics (latency, hit rate, and
retrieved-node diversity);
recent outcomes (the latest operator choices and utilities,
together with recent observed scores $y_j$);
and structural-deficiency summaries (high-degree nodes,
disconnected components, and long paths).
Episode-score features use $y_j$, not $F(\mathcal{G})$.

\subsection{Cognitive Operator Library}
\label{sec:operators}

The fixed library
$\mathcal{O}=
\{\mathcal{O}_{\mathrm{abs}},
  \mathcal{O}_{\mathrm{ref}},
  \mathcal{O}_{\mathrm{brg}},
  \mathcal{O}_{\mathrm{refr}}\}$
contains four semantic transformations.
EvoDisco learns their selection and instantiation,
not additional operator categories.

\paragraph{Abstract: consolidate related memories.}
Given a seed neighborhood, $\mathcal{O}_{\mathrm{abs}}$
synthesizes a higher-level concept or reusable procedure
and connects it to supporting memories through
\texttt{generalizes} edges.
A configured pruning criterion optionally removes
redundant memories.

\paragraph{Refine: separate overloaded memories.}
$\mathcal{O}_{\mathrm{ref}}$ decomposes a broad memory
into more selective units along a temporal, causal,
or categorical dimension.
The original node is retained as a summary or replaced,
depending on the instantiation.

\paragraph{Bridge: connect separated knowledge.}
$\mathcal{O}_{\mathrm{brg}}$ identifies semantically related
but weakly connected memory regions and generates
typed relations between them.

\paragraph{Reframe: reorient edges on a path.}
$\mathcal{O}_{\mathrm{refr}}$ keeps every node's text and
adds no node.
It rewrites \texttt{causes} and \texttt{precedes} edges
along one existing path.
The policy selects the frame, and one frozen LLM call
returns a reorientation; the executor then reverses or
relabels each consecutive pair
(Appendix~\ref{sec:appendix_operators}).

The executor proposes at most eight candidates for each
operator by a fixed rule, and masks any operator whose set
is empty.
\textsc{Abstract} ranks seeds whose $2$-hop neighborhood
contains at least two nodes, by neighborhood size and
retrieval count.
\textsc{Refine} ranks hubs by degree.
\textsc{Bridge} ranks component pairs at graph distance at
least $2$ by embedding similarity; if none exists and the
graph has at least six nodes, it proposes one pair from a
two-means split of the embeddings.
\textsc{Reframe} uses a fixed sample of long paths.
A second head selects the candidate index by a dot product
between candidate features and a learned projection of $s$.

Five categorical heads are sampled from $s$ on every attempt,
including heads the chosen operator does not read.
Neighborhood radius $k\in\{1,2,3\}$ and pruning level
$p\in\{0,0.5\}$ are read by \textsc{Abstract}, and
\textsc{Bridge} also reads $p$.
Child count $m\in\{2,3,4\}$ and a dimension in
$\{\text{temporal},\text{causal},\text{categorical}\}$
are read by \textsc{Refine}.
A logical frame in
$\{\texttt{temporal\_to\_causal},\texttt{causal\_to\_temporal},
\texttt{effect\_cause}\}$
is read by \textsc{Reframe}.
The frozen LLM writes new node text for
\textsc{Abstract}, \textsc{Refine}, and \textsc{Bridge}
only after $o$, $c$, and $\theta$ are fixed.
The \textsc{Reframe} call returns a reorientation and
writes no node.
Similarity cutoffs controlled by $p$ are fixed rules
(Appendix~\ref{sec:appendix_operators}).

\subsection{Bi-Level Fitness Framework}
\label{sec:fitness}

Feedback operates at two timescales:
immediate utility evaluates individual transformations,
and strategic efficacy guides policy updates.
Here, ``bi-level'' denotes these feedback timescales,
not an exact nested optimization procedure.

\subsubsection{Immediate Utility}
\label{sec:inner_fitness}

For $\mathcal{G}_i\rightarrow\mathcal{G}_i^{+}$, define
\begin{equation}
    U_i
    =
    \alpha\,\widehat{\Delta F}_i
    -
    \beta
    \left[
        C(\mathcal{G}_i^{+})-C(\mathcal{G}_i)
    \right],
    \label{eq:imm_util}
\end{equation}
where $\alpha,\beta>0$.
The proxy is the retrieval-quality change on queries
already observed before the attempt,
\begin{equation}
    \widehat{\Delta F}_i
    =
    \Delta Q_i
    =
    Q(\mathcal{G}_i^{+};\mathcal{Q}_i)
    -
    Q(\mathcal{G}_i;\mathcal{Q}_i),
    \label{eq:retrieval_proxy}
\end{equation}
Let $\mathcal{Q}_i^{+}$ be the cached queries that activate
at least one node, and write
$\mathrm{Top}_3(\mathcal{G},\mathbf{q})$
for the three largest spreading-activation scores.
Retrieval quality is
\begin{equation}
    Q(\mathcal{G};\mathcal{Q}_i)
    =
    \frac{1}{|\mathcal{Q}_i^{+}|}
    \sum_{\mathbf{q}\in\mathcal{Q}_i^{+}}
    \mathrm{mean}\bigl(\mathrm{Top}_3(\mathcal{G},\mathbf{q})\bigr),
    \label{eq:retrieval_quality}
\end{equation}
and $Q=0$ when $\mathcal{Q}_i^{+}$ is empty.
Both graphs are scored on the same cache immediately
after the edit, so $U_i$ does not use later episode scores.
$\widehat{\Delta F}_i$ is this graph-local surrogate.
It is not an estimate of
$F(\mathcal{G}_i^{+})-F(\mathcal{G}_i)$.
Structural complexity is
\begin{equation}
    C(\mathcal{G})
    =
    \boldsymbol{\lambda}^{\top}
    \mathbf{c}(\mathcal{G}),
    \label{eq:graph_complexity}
\end{equation}
where $\mathbf{c}(\mathcal{G})$ contains node-count,
edge-density, and path-length statistics.
An attempt that commits no edit records $U_i=-0.05$
and does not apply Eq.~\ref{eq:imm_util}.
The query cache is specified in
Appendix~\ref{sec:appendix_feedback}.

Utility is an ex post learning signal:
the transformed graph is retained regardless of the sign
of $U_i$, with no utility-based rollback.
Recorded utilities inform subsequent states, strategic
efficacy, and utility-predictor training.

\subsubsection{Strategic Efficacy}
\label{sec:outer_fitness}

Before executing $a_i$, the predictor records
$\hat{U}_i=q_{\psi}(s_i,a_i)$.
Over the most recent $m_i=\min(K,i)$ attempts,
collected in $\mathcal{W}_i$, we compute
\begin{equation}
    E_i
    =
    -\gamma\,\widehat{\mathrm{Regret}}_i
    -\delta\,\mathrm{PredErr}_i
    +\eta\,\mathrm{Div}_i.
    \label{eq:strat_eff}
\end{equation}
For each operator, $\bar{U}_i(o)$ is the mean utility of
its most recent $h=8$ executions, or of every execution
when fewer are available.
Let $\mathcal{O}_i^{\mathrm{obs}}$ be the operators with a
nonempty history, and write $[x]_{+}=\max(x,0)$.
With empirical frequency
$p_i(o)=m_i^{-1}
\sum_{\ell\in\mathcal{W}_i}\mathbb{I}[o_{\ell}=o]$,
\begin{align}
    \widehat{\mathrm{Regret}}_i
    &=
    \frac{1}{m_i}
    \sum_{\ell\in\mathcal{W}_i}
    \Big[
        \max_{o\in\mathcal{O}_i^{\mathrm{obs}}}
        \bar{U}_i(o)
        -
        U_{\ell}
    \Big]_{+},
    \label{eq:regret_proxy}\\
    \mathrm{PredErr}_i
    &=
    \frac{1}{m_i}
    \sum_{\ell\in\mathcal{W}_i}
    |\hat{U}_{\ell}-U_{\ell}|,
    \label{eq:prediction_error}\\
    \mathrm{Div}_i
    &=
    -\sum_{o\in\mathcal{O}}p_i(o)\log p_i(o),
    \label{eq:operator_diversity}
\end{align}
using $0\log 0=0$.
$\bar{U}_i(o)$ is a truncated arithmetic mean.
The positive part is zero when realized utility exceeds
that reference.
Because the reference mixes different states and
instantiations, $\widehat{\mathrm{Regret}}_i$ is not
same-state counterfactual regret.
$\mathrm{Div}_i$ is a soft preference for diversity,
not a requirement of equal usage.
The window includes attempts that do not commit an edit.

\subsection{Strategy Policy Learning}
\label{sec:policy}

A two-layer MLP parameterized by $\phi$ defines
\begin{equation}
    \pi_{\phi}(a\mid s)
    =
    \pi_{\phi}(o\mid s)\,
    \pi_{\phi}(c\mid s,o)\,
    \pi_{\phi}(\theta\mid s),
    \label{eq:joint_policy}
\end{equation}
where $o$ is the operator, $c$ is one candidate from the
executor-built set for $o$, and $\theta$ collects the
five instantiation heads.
Executor construction rules and frozen-LLM generations do not
enter the log-probability.
The weighted log-probability used for learning is
\begin{equation}
    \log\pi^{\mathrm{w}}_{\phi}(a\mid s)
    =
    \log\pi_{\phi}(o\mid s)
    +
    \log\pi_{\phi}(c\mid s,o)
    +
    \tfrac{1}{4}
    \sum_{j}
    \log\pi_{\phi}(\theta_{j}\mid s),
    \label{eq:weighted_logprob}
\end{equation}
The factor $1/4$ is a fixed scale on those heads
and is outside each categorical normalizer.

Every $N$ attempts, the policy receives a score-function
update using strategic efficacy as a window-level return:
\begin{equation}
    \widehat{\nabla_{\phi}J}_i
    =
    \operatorname{sg}(E_i)
    \sum_{\ell\in\mathcal{B}_i}
    \nabla_{\phi}
    \log\pi^{\mathrm{w}}_{\phi}(a_{\ell}\mid s_{\ell}),
    \label{eq:policy_grad}
\end{equation}
where $\mathcal{B}_i$ contains new events since the preceding
policy update and $\operatorname{sg}$ treats the observed
feedback as fixed.
The last-$K$ scoring window and the update batch have
distinct roles: older events used to calculate $E_i$
are not reused as fresh on-policy samples.

A utility predictor $q_{\psi}(s_i,a_i)$ is supervised
by recorded $U_i$ values.
State--action pairs with $U_i>0$ enter a replay buffer
$\mathcal{D}_{+}$.
Scheduled updates minimize
\begin{equation}
    \mathcal{L}_{\mathrm{replay}}(\phi)
    =
    -
    \mathbb{E}_{(s,a)\sim\mathcal{D}_{+}}
    \left[
        \log\pi^{\mathrm{w}}_{\phi}(a\mid s)
    \right].
    \label{eq:replay_loss}
\end{equation}
This auxiliary objective is separate from the
policy-gradient update; eligibility depends on utility,
not task success alone.

\subsection{Online Learning and Evaluation}
\label{sec:algorithm}

The agent processes the online stream once, writing memories
after each episode.
Evolution is enabled only during the first $\rho=0.8$
fraction of episodes.
Within this interval, an attempt is triggered either
after $T$ episodes without evolution or by a moving-average
score decline exceeding $\epsilon$
(Appendix~\ref{sec:appendix_trigger}).
Both triggers are disabled after the cutoff, while ordinary
memory writes continue.
Algorithm~\ref{alg:evodisco} in the appendix summarizes
the procedure.

For held-out evaluation, the accumulated memory is frozen:
retrieval remains active, but writes and evolution are disabled.
This frozen-memory evaluation differs from cross-benchmark
policy transfer, where policy parameters are frozen while
a new target-domain memory is constructed online.
% ============================================================
% 4. EXPERIMENTS
% ============================================================
\section{Experiments}
\label{sec:experiments}

Our experiments examine whether adaptive memory evolution improves online task performance, whether the resulting memory remains useful on held-out tasks, and whether the learned operator-selection policy transfers across environments. We evaluate EvoDisco against six baselines on four interactive benchmarks, followed by component ablations, cross-benchmark policy transfer, and an analysis of computational cost. Representative graph edits are reported in Appendix~\ref{sec:appendix_qualitative}.

\subsection{Experimental Setup}
\label{sec:setup}

\paragraph{Baselines \& Benchmarks.}
We compare EvoDisco against RAG-based (RAG, Self-RAG), agentic-memory (Mem0, MemP), and MemRL baselines, together with a no-memory agent, under a frozen-backbone setting (Appendix~\ref{sec:appendix_baselines}).
Evaluations span four benchmarks: ALFWorld (household interaction), WebShop (web shopping), LifelongAgentBench-OS (system administration), and LifelongAgentBench-DB (database manipulation).
Details are in Appendix~\ref{sec:appendix_benchmarks}.
Every method uses the same frozen backbone, GPT-5-mini; implementation choices are in Appendix~\ref{sec:appendix_impl}.

\paragraph{Metrics.}
We report success rate (SR), the fraction of episodes completed successfully.
Online SR is this ratio over the training stream, and transfer SR is the same ratio on the held-out split.

We evaluate EvoDisco and the baselines under two settings.
\textbf{Runtime Learning} assesses the ability to learn and adapt within a single pass over the training stream.
\textbf{Transferring} evaluates whether the learned memory generalizes to held-out tasks from the same benchmark once writes and evolution are disabled.
The protocol is specified in Appendix~\ref{sec:appendix_protocol}.

\subsection{Main Results}
\label{sec:main_results}

Table~\ref{tab:main} reports full-stream online SR and frozen-memory transfer SR. Cumulative online trajectories and success rates over the final $20\%$ are reported in Appendix~\ref{sec:appendix_last20}. All results are expressed as percentages; differences discussed below are in percentage points.

\begin{table}[t]
    \centering
    \caption{
        Online and frozen-memory transfer success rates (\%).
        Online SR is measured over the full training stream.
        Transfer SR is measured on the held-out split with all memory
        writes and evolution disabled.
        Bold denotes the highest value in each column.
    }
    \label{tab:main}
    \small
    \setlength{\tabcolsep}{3pt}
    \renewcommand{\arraystretch}{1.08}
    \begin{tabular*}{\linewidth}
        {@{\extracolsep{\fill}}l*{8}{c}@{}}
        \toprule
        & \multicolumn{2}{c}{ALFWorld}
        & \multicolumn{2}{c}{WebShop}
        & \multicolumn{2}{c}{OS}
        & \multicolumn{2}{c}{DB} \\
        \cmidrule(lr){2-3}
        \cmidrule(lr){4-5}
        \cmidrule(lr){6-7}
        \cmidrule(lr){8-9}
        Method
        & Online & Transfer
        & Online & Transfer
        & Online & Transfer
        & Online & Transfer \\
        \midrule
        No Memory
        & 56.7 & 52.9 & 32.2 & 32.5
        & 67.7 & 72.7 & 88.1 & 84.9 \\
        RAG
        & 60.8 & 70.0 & 29.8 & 33.0
        & 70.9 & 72.0 & 80.9 & 72.7 \\
        Self-RAG
        & 56.0 & 62.9 & 31.8 & 32.5
        & 69.7 & 74.7 & 85.3 & 84.2 \\
        Mem0
        & 58.0 & 57.9 & 29.6 & 28.0
        & 67.7 & 71.3 & 85.9 & 86.3 \\
        MemP
        & 68.9 & 75.7 & 31.4 & 34.0
        & 71.7 & 76.0 & 83.1 & 73.4 \\
        MemRL
        & 61.3 & 70.7 & 31.2 & 33.0
        & 68.6 & 72.7 & 83.7 & 74.1 \\
        \midrule
        \textbf{EvoDisco}
        & \textbf{77.7} & \textbf{88.6}
        & \textbf{32.8} & \textbf{35.0}
        & \textbf{74.0} & \textbf{77.3}
        & \textbf{88.4} & \textbf{87.1} \\
        \bottomrule
    \end{tabular*}
\end{table}

EvoDisco achieves the highest observed online and frozen-transfer SR on all four benchmarks. The largest gains occur on ALFWorld, where it reaches $77.7\%$ online SR and $88.6\%$ transfer SR, exceeding the strongest baseline, MemP, by $8.8$ and $12.9$ points, respectively. Relative to MemRL, the corresponding gains are $16.4$ and $17.9$ points. On OS, EvoDisco improves over MemP from $71.7\%$ to $74.0\%$ online and from $76.0\%$ to $77.3\%$ in transfer. The gains are smaller on WebShop and DB: transfer SR exceeds the strongest baseline by $1.0$ and $0.8$ points, respectively. These smaller margins should be interpreted as numerical differences rather than evidence of statistical significance.

The benefit of memory varies substantially across environments. On DB, several memory-based baselines underperform No Memory, whereas EvoDisco reaches $88.4\%$ online SR and $87.1\%$ transfer SR, compared with $88.1\%$ and $84.9\%$ without memory. Adding memory therefore does not uniformly improve performance in this setting.

EvoDisco also achieves the highest late-stream SR on ALFWorld, WebShop, and OS, and ties No Memory on DB (Appendix Table~\ref{tab:last20}). Appendix Figure~\ref{fig:online_sr} shows that the ALFWorld lead is present throughout training, rather than only in the final aggregate. EvoDisco's final-$20\%$ SR exceeds its full-stream average on ALFWorld ($82.4\%$ versus $77.7\%$), WebShop ($34.0\%$ versus $32.8\%$), and OS ($80.0\%$ versus $74.0\%$). On WebShop, it is the only evaluated method with a late-stream SR above its overall online SR. Because evolution is disabled during this interval, these results are consistent with continued utility of the evolved memory alongside ordinary writes. However, the comparison involves different subsets of tasks and does not by itself isolate the effect of memory growth.

\subsection{Ablation Studies}
\label{sec:ablation}

We ablate immediate utility, strategic efficacy, and the cognitive operator library, and replace learned operator selection with each of the four fixed-operator policies (Appendix~\ref{sec:appendix_component_ablation}). Table~\ref{tab:ablation} reports ALFWorld, where Full matches the main run in Table~\ref{tab:main}. OS ablations, operator-usage statistics, and strategic-efficacy trajectories are in Appendices~\ref{sec:appendix_os_ablation}, \ref{sec:appendix_operator_usage}, and \ref{sec:appendix_strategic_efficacy}.

\begin{table}[t]
    \centering
    \caption{
        ALFWorld ablations.
        Online and Transfer denote success rates (\%).
        $\Delta$Transfer is the difference from Full in percentage points.
        Bold denotes the highest success rate in each column.
    }
    \label{tab:ablation}
    \small
    \setlength{\tabcolsep}{5pt}
    \renewcommand{\arraystretch}{1.08}
    \begin{tabular*}{\linewidth}
        {@{\extracolsep{\fill}}lccc@{}}
        \toprule
        Variant & Online & Transfer & $\Delta$Transfer \\
        \midrule
        EvoDisco (Full)
        & \textbf{77.7} & \textbf{88.6} & --- \\
        \midrule
        w/o Outer Fitness       & 75.3 & 80.7 & $-7.9$ \\
        w/o Inner Fitness       & 76.0 & 79.3 & $-9.3$ \\
        w/o Cognitive Operators & 76.3 & 82.1 & $-6.5$ \\
        \midrule
        Fixed Abstract & 76.0 & 85.7 & $-2.9$ \\
        Fixed Refine   & 75.2 & 81.4 & $-7.2$ \\
        Fixed Bridge   & 75.9 & 80.7 & $-7.9$ \\
        Fixed Reframe  & 76.0 & 80.0 & $-8.6$ \\
        \bottomrule
    \end{tabular*}
\end{table}

Removing either fitness signal reduces both online and transfer performance, with substantially larger differences in frozen transfer. Without outer fitness, transfer SR decreases from $88.6\%$ to $80.7\%$; without inner fitness, it decreases to $79.3\%$. The corresponding online decreases are $2.4$ and $1.7$ points. This pattern suggests that the two feedback signals contribute not only to performance during memory construction but also to the subsequent usefulness of the resulting graph. Removing the cognitive operators likewise reduces transfer SR to $82.1\%$, a $6.5$-point decrease from Full.

Among the fixed-operator variants, \textsc{Abstract} performs best on ALFWorld, reaching $85.7\%$ transfer SR, but remains $2.9$ points below Full. Fixed \textsc{Refine}, \textsc{Bridge}, and \textsc{Reframe} trail Full by $7.2$, $7.9$, and $8.6$ points, respectively. The ordering differs on OS: \textsc{Refine} is the strongest fixed operator online at $72.9\%$, whereas \textsc{Bridge} is strongest in transfer at $76.0\%$, compared with $77.3\%$ for Full. Appendix~\ref{sec:appendix_operator_usage} does not reproduce those rankings as commit frequencies. On the logged runs, \textsc{Reframe} is the most frequent committed edit on both benchmarks, and \textsc{Abstract} commits least often because later attempts hit the abstraction cap. All four operators still occur at least once. The fixed-operator gaps and the commit frequencies are therefore different measurements.

\subsection{Analysis}
\label{sec:analysis}

\subsubsection{Cross-Benchmark Policy Transfer}

Frozen-memory evaluation tests whether a graph transfers to held-out tasks within its original benchmark. We separately examine whether the operator-selection policy $\pi$ transfers across benchmarks. Specifically, we initialize the target agent with the MLP weights learned on ALFWorld and keep those weights fixed throughout the target run. Each target memory graph is constructed from scratch using the same one-pass online and frozen-memory evaluation protocol, with no further REINFORCE updates. This experiment therefore transfers the policy, not the ALFWorld memory graph, and still permits memory construction from target training episodes.

\begin{table}[t]
    \centering
    \caption{
        Cross-benchmark transfer of the operator-selection policy.
        Values are frozen-memory transfer success rates (\%).
        The ALFWorld-trained policy is held fixed while each target
        graph is constructed from scratch.
        $\Delta$Transfer is relative to the target-trained Full model,
        in percentage points.
    }
    \label{tab:transfer}
    \small
    \setlength{\tabcolsep}{5pt}
    \renewcommand{\arraystretch}{1.08}
    \begin{tabular*}{\linewidth}
        {@{\extracolsep{\fill}}lccc@{}}
        \toprule
        Policy Training & WebShop & OS & DB \\
        \midrule
        ALFWorld              & 31.0 & 75.3 & 82.7 \\
        Target Benchmark (Full) & 35.0 & 77.3 & 87.1 \\
        \midrule
        $\Delta$Transfer & $-4.0$ & $-2.0$ & $-4.4$ \\
        \bottomrule
    \end{tabular*}
\end{table}

As shown in Table~\ref{tab:transfer}, reusing the ALFWorld-trained policy yields $31.0\%$, $75.3\%$, and $82.7\%$ transfer SR on WebShop, OS, and DB, which are $4.0$, $2.0$, and $4.4$ points below the target-trained models. Appendix~\ref{sec:appendix_operator_usage} is a separate commit log, not the transferred checkpoint, so the gaps should not be read off those frequencies. These results do not establish that the source-trained policy is preferable to an untrained selector; that comparison would require an additional control.

\subsubsection{Computational Cost}

An EvoDisco evolution step consists of local graph-statistics computation and state encoding, a forward pass through the policy network, and an LLM call when the selected operator requires generation or relation inference. Immediate utility is computed locally from retrieval quality and graph complexity; the default implementation does not invoke a separate LLM call to estimate operator effects. Thus, an evolution event requires approximately one operator-generation call when applicable, in addition to local computation, rather than evaluating a population of graph variants through repeated task rollouts.

In the reported ALFWorld runs, end-to-end wall-clock time is $21.6$ hours for EvoDisco, compared with $33.6$ hours for MemRL and $36.8$ hours for No Memory. On OS, EvoDisco completes in $4.2$ hours, compared with $6.1$ hours for MemRL. These measurements include task execution and memory maintenance: successful episodes can end early, while unsuccessful ALFWorld episodes may reach the $50$-step cap, so the lower runtime is total agent cost rather than isolated evolution overhead.
% ============================================================
% 5. CONCLUSION
% ============================================================
\section{Conclusion}
\label{sec:conclusion}

EvoDisco learns which cognitive operator to apply when evolving agent memory, instead of following a hand-specified rule. Its policy selects among \textsc{Abstract}, \textsc{Refine}, \textsc{Bridge}, and \textsc{Reframe} and is trained with bi-level feedback: retrieval-quality gain regularized by structural complexity, together with operator-utility regret, utility-prediction error, and selection diversity. The study is limited by a fixed library of four operators, a trigger that is scheduled or heuristic rather than learned, and evaluation with a single backbone and an ALFWorld-only transfer source. Within that scope, EvoDisco matches or exceeds MemP and MemRL on ALFWorld, WebShop, and LifelongAgentBench (OS and DB). The largest gain is on ALFWorld, where online and frozen-transfer success rise by $8.8$ and $12.9$ points over MemP, and ablations tie that transfer gap to both fitness signals and learned operator selection.

\bibliographystyle{unsrtnat}
\bibliography{evodisco}

\appendix
% ============================================================
% 6. APPENDIX
% ============================================================
\section{Additional Implementation Details}
\label{sec:appendix}

This appendix provides the experimental protocol, benchmarks, baselines, and implementation choices deferred from Section~\ref{sec:setup}, together with further implementation details for EvoDisco, including the shared hyperparameter setting, operator instantiation settings, the evolution-trigger configuration, the execution of the three component ablations, OS ablations (Table~\ref{tab:ablation_os}), operator-usage statistics, strategic-efficacy trajectories, two qualitative graph edits (Figure~\ref{fig:qualitative}), and the shared prompt templates (Appendix~\ref{sec:appendix_prompts}).

\subsection{Evaluation Protocol}
\label{sec:appendix_protocol}

We adopt a \emph{one-pass online learning followed by frozen-memory transfer} protocol, with the backbone LLM held fixed throughout.
Runtime Learning is the online phase: each agent processes the training split once and may write memories after every episode.
For methods with memory evolution, evolution is restricted to the first $80\%$ of the training stream.
Ordinary memory writes remain enabled throughout the stream, including the final $20\%$.

We report success rate (SR) over the full online stream and separately over its final $20\%$.
The latter measures late-stream performance after evolution has stopped, although ordinary memory accumulation continues; those rates are in Appendix~\ref{sec:appendix_last20}.
Transferring is the frozen-memory phase.
After the online pass, we freeze all memory state and evaluate on the held-out split without further writes or evolution.
Frozen-memory transfer is our primary measure of whether the accumulated memory supports unseen tasks within the same benchmark.

\subsection{Benchmarks}
\label{sec:appendix_benchmarks}

We evaluate on four benchmarks chosen to cover household interaction, web shopping, operating-system administration, and database manipulation.
Within each benchmark, every method uses the same agent loop, split files, backbone LLM, and step budget.
Split sizes and budgets are summarized in Table~\ref{tab:splits}.

\begin{itemize}
    \item \textbf{ALFWorld}~\citep{shridhar2021alfworld}:
    A text-based household environment for embodied navigation and manipulation.
    Each episode states a goal in natural language, such as finding an object and placing it at a destination, and the agent must issue a sequence of high-level actions (navigate, open, take, and place) until the goal holds or the step limit is reached.
    Success therefore depends on retrieving a multi-step plan that matches the current scene, rather than on a single lookup.
    We use all $3553$ training episodes and the $140$-task in-distribution evaluation split, with a budget of $50$ steps per episode.

    \item \textbf{WebShop}~\citep{yao2022webshop}:
    A simulated shopping website on which the agent must satisfy a natural-language product instruction by searching, selecting an item, and choosing options.
    The environment returns a page observation at each step, so the agent has to keep the instruction constraints in view while navigating search results.
    In addition to binary success, WebShop assigns a graded purchase score that reflects how closely the chosen product matches the instruction.
    We use $500$ training instructions and $200$ validation instructions, with a budget of $15$ steps per episode.
    Reported tables use binary success.

    \item \textbf{LifelongAgentBench-OS}~\citep{zheng2025lifelongagentbench}:
    Interactive Ubuntu/Bash administration tasks.
    The agent issues shell commands, reads their output, and must finish the requested administration steps over a multi-turn session.
    The benchmark is designed for continual tool use: later tasks reuse commands and patterns that appeared in earlier tasks.
    We use the official MemRL split of $350$ training tasks and $150$ validation tasks, with a budget of $15$ steps per episode.

    \item \textbf{LifelongAgentBench-DB}~\citep{zheng2025lifelongagentbench}:
    SQL interaction with a live database.
    The agent writes queries, inspects results, and revises them until the requested table state or answer is obtained.
    Tasks share a schema, so a query pattern learned on one task can apply to a later request against the same tables.
    We use the official split of $361$ training tasks and $139$ validation tasks, with a budget of $15$ steps per episode.
\end{itemize}

\subsection{Baselines}
\label{sec:appendix_baselines}

We compare against protocol-matched reimplementations of the memory systems in the MemRL evaluation~\citep{zhang2026memrl}.
Every baseline uses the same frozen backbone, agent loop, splits, and step budgets as EvoDisco.
During Runtime Learning a baseline may write after each episode; during Transferring its memory is frozen.
None of these systems selects or instantiates a graph-evolution operator.

\begin{itemize}
    \item \textbf{No Memory}:
    The agent conditions only on the current episode: the task text, the latest observation, and the actions already taken in that episode.
    It does not write a note, embed past experience, or retrieve anything from previous episodes.
    This baseline measures task performance when no cross-episode memory is available.

    \item \textbf{RAG}~\citep{lewis2020retrieval}:
    After each episode the backbone writes a reflection of the trajectory and stores it as an embedding.
    At the start of a new episode the system embeds the task and returns the top-$k$ notes by cosine similarity.
    Retrieval ranks notes by cosine similarity alone.
    Both successful and failed episodes are stored, so later retrieval can surface either kind of note.

    \item \textbf{Self-RAG}~\citep{asai2024selfrag}:
    Self-RAG uses the same note store as RAG and adds a decide-then-critique filter.
    Before retrieval, the backbone answers whether stored memories would help the current task; a negative answer skips retrieval.
    When retrieval runs, each candidate is marked keep or drop, and only the kept notes are passed to the agent.
    The filter is intended to suppress notes that are similar in embedding space but not useful for the current instruction.

    \item \textbf{Mem0}~\citep{chhikara2025mem0}:
    A memory layer that maintains a persistent store through add, retrieve, and update operations.
    After an episode it extracts one to three atomic facts---a lesson, a constraint, or a working pattern---rather than saving the raw trajectory.
    Each fact is embedded and compared with existing notes.
    If the nearest note has cosine similarity of at least $0.82$, that note is overwritten with the new fact; otherwise the fact is inserted as a new note.
    Retrieval is the same cosine top-$k$ used by RAG.

    \item \textbf{MemP}~\citep{fang2025memp}:
    A procedural-memory method.
    On a successful episode it distills the trajectory into a numbered, high-level procedure that states the sequence of operations that worked, and stores that procedure for later retrieval.
    The stored text is this distilled procedure.
    On a failed episode it does not add a new procedure; it instead lowers the reliability of the procedures that were retrieved for that episode.
    The resulting store is updated by this build--retrieve--update cycle and is queried by cosine similarity.

    \item \textbf{MemRL}~\citep{zhang2026memrl}:
    An episodic-memory baseline that keeps the backbone frozen and ranks stored notes with a two-phase score.
    The first phase retrieves candidates by semantic similarity; the second re-ranks them with a non-parametric $Q$-value estimated from past task outcomes.
    The memory topology is held fixed: MemRL updates the utility used at retrieval, and it does not rewrite, merge, or restructure the stored notes.
    We use the same frozen-backbone protocol as the other baselines.

    \item \textbf{Fixed-Strategy Evolution}:
    This variant is an ablation of EvoDisco.
    Each variant clamps the operator to one of \textsc{Abstract}, \textsc{Refine}, \textsc{Bridge}, or \textsc{Reframe} and applies that operator at every evolution step.
    The candidate and the instantiation parameters are still sampled, but the policy receives no REINFORCE update, so the choice of operator does not change with experience.
    These four runs are reported in Section~\ref{sec:ablation} and Appendix~\ref{sec:appendix_os_ablation}.
\end{itemize}

\subsection{Implementation Details}
\label{sec:appendix_impl}

All agents use \textbf{GPT-5-mini} with temperature $0.0$ and \texttt{text-embedding-3-large} for embeddings.
Memory graphs are maintained in process rather than in an external graph database.
The strategy policy $\pi$ is a two-layer MLP with hidden dimension $128$, optimized using REINFORCE-style updates with a learning rate of $1\times 10^{-3}$.

Immediate utility weights the online performance proxy $\widehat{\Delta F}=\Delta Q$ and the structural-complexity change $\Delta C$ by $\alpha=1.0$ and $\beta=0.3$.
Strategic efficacy uses $\gamma=1.0$, $\delta=1.0$, and $\eta=0.3$, with a history window of $K=20$ and a policy-update interval of $N=5$.
Within the evolution window, a hybrid trigger initiates evolution every $T=10$ episodes or when the moving average of the last $10$ episode scores decreases by more than $\epsilon=0.15$.
Neither trigger permits evolution beyond the first $80\%$ of the online stream.

At each evolution attempt, the policy samples an operator, one executor-built candidate, and the instantiation parameters under $\pi$.
An attempt is counted as an applied operation only when that choice commits a graph edit.
Retrieval, memory-writing formats, and domain routing follow the per-benchmark configuration files released with the code.
The shared hyperparameter setting is in Table~\ref{tab:hparams}, and the trigger is specified further in Appendix~\ref{sec:appendix_trigger}.

\subsection{Memory Attributes and State Encoding}
\label{sec:appendix_memory_state}

Section~\ref{sec:problem} specifies node attributes,
the relation vocabulary, and the four feature groups in $s_i$.
The lengths of the recent-outcome windows, feature normalization,
and the encoding of variable-size deficiency summaries are
left as implementation conventions.

% AUTHOR CHECK:
% Specify L, M, feature normalization, and padding conventions.
% Clarify how variable-size deficiency summaries are encoded.

\subsection{Feedback Definitions}
\label{sec:appendix_feedback}

\paragraph{Online performance proxy.}
\Eqref{eq:retrieval_quality} defines $Q$.
The implementation retains the most recent $32$ query
embeddings observed before event $i$ and scores the last $12$.
Spreading activation returns node scores
$a_v(\mathcal{G},\mathbf{q})$.
No later episode, held-out task, or task score $y_j$
enters $\Delta Q_i$.

\paragraph{Structural complexity.}
\Eqref{eq:graph_complexity} defines $C(\mathcal{G})$.
The vector $\mathbf{c}(\mathcal{G})$ uses the configured
node-count, edge-density, and path-length statistics,
and $\boldsymbol{\lambda}$ gives their relative weights.

% AUTHOR CHECK:
% Specify lambda, feature scaling, and density/path-length
% conventions for a multi-relational, disconnected graph.

\paragraph{Pre-execution utility prediction.}
The forecast $\hat{U}_i=q_{\psi}(s_i,a_i)$ is stored before
execution, as used in Eq.~\ref{eq:prediction_error}.
Stored predictions are not recomputed after predictor
updates, so each prediction error uses the forecast
available before its outcome.

\paragraph{Scoring window.}
Equations~\ref{eq:regret_proxy}--\ref{eq:operator_diversity}
are evaluated on
$\mathcal{W}_i=\{\max(1,i-K+1),\ldots,i\}$.
The operator mean $\bar{U}_i(o)$ averages the most recent
$h=8$ executions of $o$, or every execution when fewer
than $h$ are available.

\subsection{Learning and Execution Details}
\label{sec:appendix_learning}
\label{sec:dual_loop}

\paragraph{Policy decisions.}
\Eqref{eq:joint_policy} is the sampling distribution,
and Eq.~\ref{eq:weighted_logprob} is the log-probability
substituted into
Eqs.~\ref{eq:policy_grad} and~\ref{eq:replay_loss}.
Section~\ref{sec:operators} states which heads each operator
reads; Section~\ref{sec:appendix_operators} records the
numeric cutoffs.
Full runs draw $o$ from the four-operator library, with no
clamped operator and no benchmark-specific operator prior.

\paragraph{Scoring versus update batches.}
The scoring window $\mathcal{W}_i$ contains the most
recent $K$ attempts, whereas $\mathcal{B}_i$ contains
new attempts since the preceding policy update.
Only $\mathcal{B}_i$ supplies action log-probabilities
in Eq.~\eqref{eq:policy_grad}.
An older event can contribute to the aggregate score
without being treated as a fresh on-policy sample.

% AUTHOR CHECK:
% Match this convention to the implementation.
% If policy-gradient updates reuse actions collected before
% a policy change, document the treatment of stale samples.

\paragraph{Utility prediction and positive replay.}
The predictor is trained on recorded
$(s_i,a_i,U_i)$ tuples.
State--action pairs enter $\mathcal{D}_{+}$ only
when $U_i>0$, and scheduled updates minimize
Eq.~\ref{eq:replay_loss}.

\paragraph{Two feedback timescales.}
Immediate utility updates the records used for subsequent
decisions after each attempt.
Strategic efficacy supplies an aggregate policy-learning
signal every $N$ attempts.
Neither utility calculation directly changes the graph;
graph edits are performed by operator execution.

\begin{table}[t]
\centering
\caption{Two timescales of evolution feedback.
Graph modification is performed by operator execution.}
\label{tab:dual_loop}
\small
\begin{tabular}{llll}
\toprule
Level & Signal & Learning role & Frequency \\
\midrule
Inner & $U_i$ & Feedback and supervision & Each attempt \\
Outer & $E_i$ & Strategy-policy update & Every $N$ attempts \\
\bottomrule
\end{tabular}
\end{table}

\paragraph{Attempts and committed edits.}
An evolution event is a triggered attempt.
Attempts that leave the graph unchanged remain in
the event history and count toward the policy-update
interval.
Operator-usage statistics separately report committed
graph edits, as described in
Appendix~\ref{sec:appendix_operator_usage}.
No utility-based rollback is performed.

\paragraph{Overall procedure.}
Algorithm~\ref{alg:evodisco} gives the online loop.
Evolution is disabled after the first
$\lfloor\rho n\rfloor$ episodes, where $n$ is the online
stream size and $\rho=0.8$.
Ordinary writes continue until the stream ends,
after which the graph is frozen for held-out evaluation.

\begin{algorithm}[t]
\caption{EvoDisco: Strategy Learning for Memory Evolution}
\label{alg:evodisco}
\begin{algorithmic}[1]
\Require Online stream of $n$ episodes;
initial graph $\mathcal{G}_0$;
policy $\pi_{\phi}$; predictor $q_{\psi}$;
operator library $\mathcal{O}$;
scoring window $K$; update interval $N$
\State $\mathcal{G}\leftarrow\mathcal{G}_0$,
       $i\leftarrow 0$
\State $\mathcal{H}\leftarrow[\,]$,
       $\mathcal{B}\leftarrow[\,]$,
       $\mathcal{D}_{+}\leftarrow\emptyset$
\For{each online task episode $j$}
    \State Run the agent with $\mathcal{G}$;
           observe $y_j$ and interaction history
    \State Write the new experience into $\mathcal{G}$
    \If{$j\leq\lfloor\rho n\rfloor$ and trigger is met}
        \State $i\leftarrow i+1$
        \State $s_i\leftarrow
               \mathrm{encode\_state}
               (\mathcal{G},\mathcal{H},y_{1:j})$
        \State Sample $a_i=(o_i,c_i,\theta_i)
               \sim\pi_{\phi}(\cdot\mid s_i)$
        \State $\hat{U}_i\leftarrow q_{\psi}(s_i,a_i)$
        \State $\mathcal{G}^{+}\leftarrow
               \mathsf{T}_{o_i}(\mathcal{G}; c_i, \theta_i)$
        \If{the operator commits an edit}
            \State Compute $U_i$ using Eq.~\eqref{eq:imm_util}
        \Else
            \State $U_i\leftarrow -0.05$
        \EndIf
        \State $\mathcal{G}\leftarrow\mathcal{G}^{+}$
               \Comment{No utility-based rollback}
        \State Append $(s_i,a_i,\hat{U}_i,U_i)$
               to $\mathcal{H}$ and $\mathcal{B}$
        \If{$U_i>0$}
            \State Add $(s_i,a_i)$ to $\mathcal{D}_{+}$
        \EndIf
        \State Update $q_{\psi}$ using recorded utility targets
        \If{$i\bmod N=0$}
            \State Compute $E_i$ on the last $\min(K,i)$
                   attempts using Eq.~\eqref{eq:strat_eff}
            \State Update $\phi$ using
                   Eq.~\eqref{eq:policy_grad}
            \State Apply scheduled replay updates using
                   Eq.~\eqref{eq:replay_loss}
            \State $\mathcal{B}\leftarrow[\,]$
        \EndIf
    \EndIf
\EndFor
\State Freeze $\mathcal{G}$ for held-out evaluation
\end{algorithmic}
\end{algorithm}

\subsection{Policy-Controlled Parameters}
\label{sec:appendix_operators}

Section~\ref{sec:operators} gives the candidate-construction
rules and the supports of the five instantiation heads.
The executor does not draw another target after $c$ is sampled.
\textsc{Abstract} retrieves the $k$-hop neighborhood of the chosen seed and prunes a supporting node when $p>0$, its similarity to the new concept is at least $0.70+0.25p$, and it fails the fixed retention checks on retrieval count and degree.
\textsc{Bridge} has no separate threshold head: it rejects the chosen pair when the embedding similarity is below $0.35+0.3p$.
The distance cutoff in Section~\ref{sec:operators} is a minimum: candidate pairs stand at graph distance at least $2$.
That cutoff is a construction rule, not a policy output.

\textsc{Reframe} keeps every node's text, writes no node, and rewrites only \texttt{causes} and \texttt{precedes} edges along one existing path.
The sampled frame replaces any frame echoed by the model.
The executor reads \texttt{REORIENT} and ignores \texttt{SUMMARY}.
On each consecutive pair $(a,b)$ the reverse branch applies when \texttt{REORIENT} is \texttt{reverse} or the frame is \texttt{effect\_cause}.
That branch moves an existing \texttt{precedes} or \texttt{causes} edge from $a$ to $b$ onto $b$ to $a$, keeping weight $\max(w,0.4)$, and then writes \texttt{causes} from $b$ to $a$.
Otherwise \texttt{causal\_to\_temporal} writes \texttt{precedes} from $a$ to $b$, and \texttt{temporal\_to\_causal} writes \texttt{causes} from $a$ to $b$.
\texttt{relabel\_causes} and \texttt{relabel\_precedes} both take this second branch.
A missing edge is inserted at weight $0.75$.
An edge that is already present, including one just reversed, keeps its weight and adds $0.15\times 0.75$.
After that write, every \texttt{causes} or \texttt{precedes} edge on the pair, in either direction, has its weight increased by $0.1$, clipped to $[0,1]$, and its recency set to the current task index.

\paragraph{Fixed by rule or by the LLM.}
The frozen LLM writes the \textsc{Abstract} concept, the \textsc{Refine} children, the \textsc{Bridge} relation type, and the \textsc{Reframe} reorientation.
The endpoint pair inside a bridge candidate is the closest embedding pair.
For \textsc{Refine}, the relation label follows from the sampled dimension once $\theta$ is chosen.
For \textsc{Reframe}, the edit follows from the sampled frame together with whether the returned reorientation is \texttt{reverse}.
The numeric cutoffs above are fixed rules.
The prompts are the shared templates in Appendix~\ref{sec:appendix_prompts}.

\subsection{Evolution Trigger}
\label{sec:appendix_trigger}

We use a hybrid evolution trigger. An evolution step is executed when either (i) $T=10$ new task episodes have elapsed since the last evolution, or (ii) the moving average of the last $10$ episode scores drops by more than $\epsilon=0.15$. Evolution stops after $80\%$ of the online stream; frozen transfer never writes or evolves. This design prevents both overly frequent reorganization and delayed response to memory degradation.

\noindent\begin{minipage}{\textwidth}
\vspace{0.6em}
\subsection{Hyperparameters}
\label{sec:appendix_hparams}

Table~\ref{tab:hparams} lists the setting used for all four benchmarks.
We chose these values by repeated trials on a development split held out from the reported evaluation, then froze them for the online and frozen-transfer runs.
All four benchmarks share this setting.
Primary terms use unit weight ($\alpha=\gamma=\delta=1$); $\beta=\eta=0.3$ keep the structural-complexity penalty and the diversity term on a smaller scale.

\begin{table}[H]
\centering
\footnotesize
\setlength{\tabcolsep}{4pt}
\renewcommand{\arraystretch}{0.96}
\caption{Hyperparameters used in the reported EvoDisco experiments.}
\label{tab:hparams}
\begin{tabularx}{\linewidth}{@{}ll>{\raggedright\arraybackslash}X@{}}
\toprule
Symbol / Name & Value & Description \\
\midrule
$\alpha$ & $1.0$ & Immediate-utility weight on the online performance proxy $\widehat{\Delta F}=\Delta Q$ \\
$\beta$ & $0.3$ & Immediate-utility weight on structural-complexity change $\Delta C$ \\
$\gamma$ & $1.0$ & Strategic-efficacy weight on $\widehat{\mathrm{Regret}}$ \\
$\delta$ & $1.0$ & Strategic-efficacy weight on $\mathrm{PredErr}$ \\
$\eta$ & $0.3$ & Strategic-efficacy weight on operator diversity \\
$K$ & $20$ & Strategic-efficacy window size \\
$N$ & $5$ & Policy-gradient interval (evolution attempts) \\
$T$ & $10$ & Episodes between evolution triggers \\
$\epsilon$ & $0.15$ & Drop in the 10-episode moving-average score that fires evolution \\
Score window & $10$ & Episodes in the moving average for the drop trigger \\
Cutoff & $0.8$ & Fraction of online stream after which evolution stops \\
Hidden dim. & $128$ & MLP policy hidden size \\
Learning rate & $1\times 10^{-3}$ & REINFORCE step size \\
Temperature & $0.0$ & Backbone sampling temperature \\
Embedding & \texttt{text-embedding-3-large} & Text encoder \\
Seed & $42$ & Split shuffle / runner seed \\
\bottomrule
\end{tabularx}
\end{table}
\end{minipage}

\subsection{Splits and Budgets}
\label{sec:appendix_splits}

Table~\ref{tab:splits} lists the online-stream size, the frozen-transfer split size, and the per-episode step budget.

\begin{table}[H]
\centering
\small
\caption{Split sizes and per-episode step budgets.}
\label{tab:splits}
\begin{tabular}{lccc}
\toprule
Benchmark & Online $n$ & Transfer $n$ & Max steps \\
\midrule
ALFWorld & $3553$ & $140$ & $50$ \\
WebShop & $500$ & $200$ & $15$ \\
LifelongAgentBench-OS & $350$ & $150$ & $15$ \\
LifelongAgentBench-DB & $361$ & $139$ & $15$ \\
\bottomrule
\end{tabular}
\end{table}

\subsection{Component Ablation Execution}
\label{sec:appendix_component_ablation}

The rows \emph{w/o Outer Fitness}, \emph{w/o Inner Fitness}, and \emph{w/o Cognitive Operators} in Table~\ref{tab:ablation} and Table~\ref{tab:ablation_os} each remove one component of the evolution step. The online-then-frozen protocol, splits, backbone, step budgets, retrieval, ordinary memory writes, hybrid trigger, and $80\%$ evolution cutoff are those of the Full run. An attempt is applied only when the chosen edit commits a graph change.

\paragraph{Without outer fitness.}
Outer fitness is the strategic-efficacy signal $E_i$ (Section~\ref{sec:outer_fitness}) that trains the strategy policy. This variant keeps the cognitive operators and immediate utility, and disables policy selection and policy learning. At each evolution attempt the executor builds the same per-operator candidate sets as Full. It then draws an operator uniformly among those with a nonempty candidate set, draws one candidate uniformly from that set, and draws the instantiation uniformly from the grids in Section~\ref{sec:operators}: radius $k\in\{1,2,3\}$, child count $m\in\{2,3,4\}$, pruning threshold, decomposition dimension, and logical frame. The drawn operator is executed with the same generator used by Full. Immediate utility follows the committed-edit rule in Eq.~\eqref{eq:imm_util}, and a non-committed attempt stores $U_i=-0.05$. The MLP is not queried, and the REINFORCE update that would apply $E_i$ every $N$ events is skipped, so later attempts remain uniform over the valid candidates.

\paragraph{Without inner fitness.}
Inner fitness is the immediate utility $U_i$ (Section~\ref{sec:inner_fitness}). Candidate construction, policy sampling, cognitive-operator execution, and the outer-loop update every $N=5$ events match Full. The retrieval-quality change and the complexity penalty are omitted from the recorded outcome: every event stores $U_i=0$, whether or not the operator commits an edit. The utility predictor is supervised with these zero targets. The replay buffer, which admits only events with $U_i>0$, receives no events. Strategic efficacy is still evaluated on the resulting history, so $\widehat{\mathrm{Regret}}_i$ and $\mathrm{PredErr}_i$ are computed from zero utilities while $\mathrm{Div}_i$ remains the operator entropy of the window. The policy therefore continues to update, and the update no longer depends on retrieval quality or structural complexity.

\paragraph{Without cognitive operators.}
This variant keeps the hybrid trigger and replaces the library $\{\textsc{Abstract},\textsc{Refine},\textsc{Bridge},\textsc{Reframe}\}$ with three structural edits. No candidate set is built, the policy is not queried, and the language model is not called to author new node text. Each attempt draws one edit uniformly.
\begin{itemize}
    \item \textbf{Add edge.} Two distinct nodes are drawn uniformly. A relation is drawn uniformly from \texttt{analogous-to}, \texttt{causes}, \texttt{precedes}, and \texttt{is-a}, and inserted with weight $0.5$.
    \item \textbf{Delete edge.} One existing relation is drawn uniformly and removed. When the graph contains no relations, the attempt commits no edit.
    \item \textbf{Merge.} Two distinct nodes are drawn uniformly. The retained node's text is the concatenation of the two contents, truncated at $2000$ characters. Its embedding is the $\ell_2$-normalized sum of the two embeddings, and its confidence is their arithmetic mean. Relations incident on the discarded node are rewritten onto the retained node, after which the discarded node is removed.
\end{itemize}
Immediate utility is computed on the edited graph and stored. The policy receives no update, so the next edit is again a uniform draw over the three structural edits. A graph with fewer than two nodes commits no edit.

The fixed-operator rows are a separate intervention. Each clamps $o$ to one named cognitive operator. The initial MLP still samples that operator's candidate and the instantiation heads in Section~\ref{sec:operators}, and those weights receive no REINFORCE update. These rows do not use the structural edits above.

\subsection{OS Ablations}
\label{sec:appendix_os_ablation}

Table~\ref{tab:ablation_os} reports the OS analogue of Table~\ref{tab:ablation}. EvoDisco (Full) is the same system as Table~\ref{tab:main} ($74.0$ online / $77.3$ transfer).

\begin{table}[htbp]
\centering
\caption{OS ablations (\%). Deltas are relative to Full (Table~\ref{tab:main}). Best in each column is bold.}
\label{tab:ablation_os}
\small
\begin{tabular}{lccc}
\toprule
Variant & Online & Xfer & $\Delta$Xfer \\
\midrule
EvoDisco (Full) & \textbf{74.0} & \textbf{77.3} & --- \\
w/o Outer Fitness & 71.1 & 75.3 & $-2.0$ \\
w/o Inner Fitness & 69.1 & 74.0 & $-3.3$ \\
w/o Cognitive Operators & 71.7 & 74.0 & $-3.3$ \\
Fixed Abstract & 72.0 & 74.0 & $-3.3$ \\
Fixed Refine & 72.9 & 75.3 & $-2.0$ \\
Fixed Bridge & 71.7 & 76.0 & $-1.3$ \\
Fixed Reframe & 71.1 & 74.0 & $-3.3$ \\
\bottomrule
\end{tabular}
\end{table}

Removing outer fitness costs $2.0$ transfer points; removing inner fitness is the largest drop ($-3.3$ transfer and $-4.9$ online). w/o Cognitive Operators matches that transfer drop. No fixed operator beats Full on either metric; Fixed \textsc{Refine} is the strongest fixed variant on online SR ($72.9\%$), and Fixed \textsc{Bridge} is closest on frozen transfer ($76.0\%$, $-1.3$).

\section{Additional Experimental Analysis}
\label{sec:appendix_analysis}

\subsection{Late-Stream Online Success Rates}
\label{sec:appendix_last20}

\begin{figure*}[t]
    \centering
    \includegraphics[width=\textwidth]{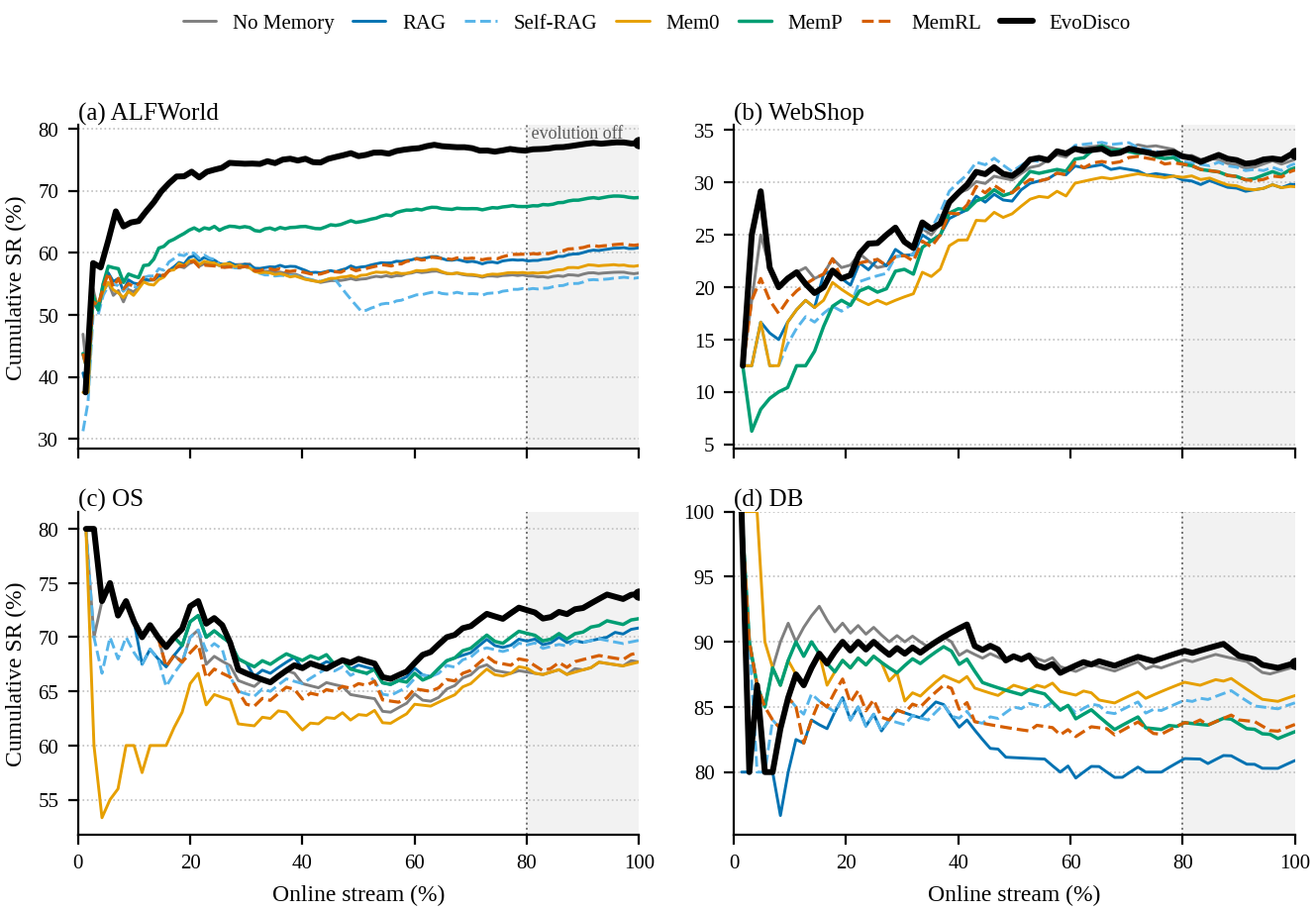}
    \caption{
        Cumulative online success rate over the training stream.
        Each point is the success rate on all episodes observed so far.
        The shaded interval is the final $20\%$, during which memory
        evolution is disabled and ordinary memory writes continue.
        EvoDisco separates from every baseline on ALFWorld and finishes
        highest on all four benchmarks.
        On WebShop and DB the curves overlap for much of the stream,
        and the final margins are small.
    }
    \label{fig:online_sr}
\end{figure*}

Table~\ref{tab:last20} reports success rates over the final $20\%$ of the online stream. Memory evolution is disabled during this interval, while ordinary memory writes remain enabled. Figure~\ref{fig:online_sr} plots cumulative online SR over the full stream. The right-hand edge of each curve is the full-stream online SR in Table~\ref{tab:main}; the shaded suffix is the interval summarized by Table~\ref{tab:last20}.

\begin{table}[htbp]
    \centering
    \caption{
        Success rates (\%) over the final $20\%$ of the online stream.
        Memory evolution is disabled during this interval, while
        ordinary memory writes remain enabled.
        Bold denotes the highest value in each column, including ties.
    }
    \label{tab:last20}
    \small
    \setlength{\tabcolsep}{5pt}
    \renewcommand{\arraystretch}{1.08}
    \begin{tabular*}{\linewidth}
        {@{\extracolsep{\fill}}lcccc@{}}
        \toprule
        Method & ALFWorld & WebShop & OS & DB \\
        \midrule
        No Memory & 58.6 & 31.0 & 71.4 & \textbf{86.1} \\
        RAG       & 69.1 & 28.0 & 75.7 & 80.6 \\
        Self-RAG  & 63.4 & 31.0 & 71.4 & 84.7 \\
        Mem0      & 62.7 & 26.0 & 70.0 & 81.9 \\
        MemP      & 74.7 & 30.0 & 77.1 & 80.6 \\
        MemRL     & 67.4 & 29.0 & 71.4 & 83.3 \\
        \midrule
        \textbf{EvoDisco}
        & \textbf{82.4} & \textbf{34.0}
        & \textbf{80.0} & \textbf{86.1} \\
        \bottomrule
    \end{tabular*}
\end{table}

On ALFWorld, EvoDisco pulls ahead early and remains clearly above MemP for the rest of the stream. On OS, EvoDisco is highest at the evolution cutoff and at the end of training, although MemP is narrowly ahead near $40\%$ progress ($67.9\%$ versus $67.1\%$). On WebShop and DB the trajectories overlap. EvoDisco still finishes first, by $0.6$ and $0.3$ points over the next-best online SR.

\subsection{Operator Usage Distribution}
\label{sec:appendix_operator_usage}

\begin{figure*}[t]
\centering
\includegraphics[width=\textwidth]{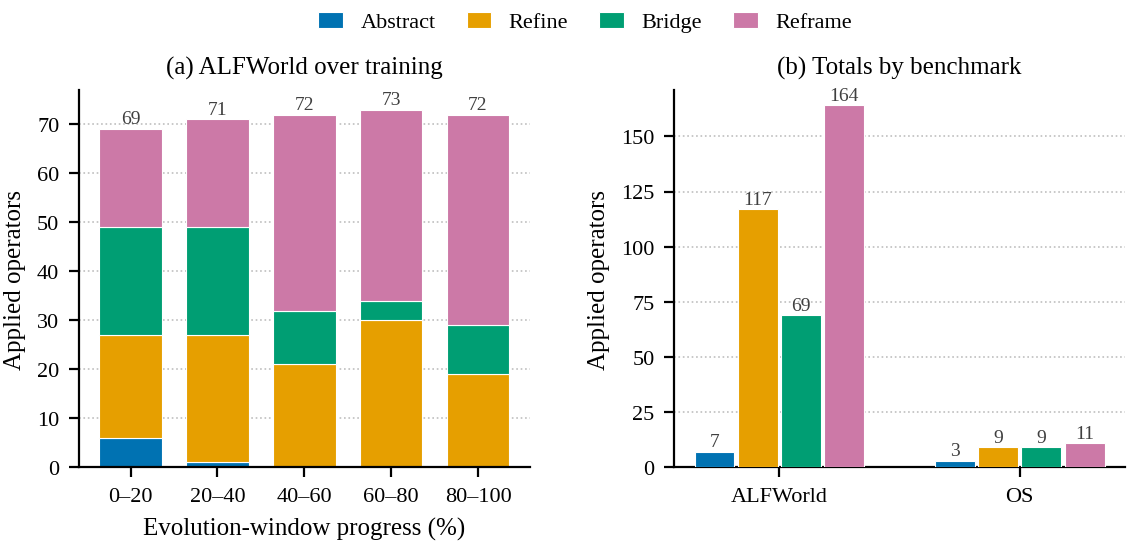}
\caption{Applied operators. (a)~ALFWorld, five equal bins of the evolution window (first $2842$ episodes). (b)~Totals on ALFWorld ($357$ commits) and OS ($32$ commits). \textsc{Abstract} is confined to the early ALFWorld bins. \textsc{Reframe} is the most frequent commit on both benchmarks.}
\label{fig:operator_dist}
\end{figure*}

Figure~\ref{fig:operator_dist} and Table~\ref{tab:operator_counts} report operators that commit a graph edit. The logs contain $400$ evolution attempts on ALFWorld, all inside the first $80\%$ of the training stream ($2842$ episodes), and $39$ attempts on OS, inside the first $280$ of $350$ training episodes. Of these, $357$ and $32$ are applied. The remainder are \textsc{Abstract} attempts that return \texttt{max\_abstraction} and leave the graph unchanged ($43$ on ALFWorld, $7$ on OS). \textsc{Refine}, \textsc{Bridge}, and \textsc{Reframe} commit on every logged attempt. Policy-gradient updates run every $N=5$ attempts, which yields $80$ updates on ALFWorld and $7$ on OS. Failed attempts stay in the strategic-efficacy window; they are omitted from Figure~\ref{fig:operator_dist} and Table~\ref{tab:operator_counts}.

\begin{table}[htbp]
\centering
\caption{Applied operator counts. Attempts that do not commit an edit are excluded. Updates are policy-gradient steps, one every $N=5$ attempts.}
\label{tab:operator_counts}
\small
\begin{tabular}{lcccccc}
\toprule
Benchmark & Abstract & Refine & Bridge & Reframe & Total & Updates \\
\midrule
ALFWorld & $7$ & $117$ & $69$ & $164$ & $357$ & $80$ \\
OS & $3$ & $9$ & $9$ & $11$ & $32$ & $7$ \\
\bottomrule
\end{tabular}
\end{table}

The mix is not uniform, and it does not collapse to one operator:
\begin{itemize}
    \item \textbf{ALFWorld.} The first fifth of the evolution window still applies all four operators ($6$ \textsc{Abstract}, $21$ \textsc{Refine}, $22$ \textsc{Bridge}, $20$ \textsc{Reframe}). \textsc{Abstract} then drops out: $6$ of its $7$ committed edits fall in that first bin, and none are committed after $40\%$ of the window, because later attempts hit the abstraction cap. \textsc{Reframe} rises from $20$ to $43$ commits per bin. \textsc{Refine} stays between $19$ and $30$. \textsc{Bridge} falls from $22$ to $4$ and then returns to $10$ in the last bin. The window totals are $164$ \textsc{Reframe}, $117$ \textsc{Refine}, $69$ \textsc{Bridge}, and $7$ \textsc{Abstract}.
    \item \textbf{OS.} The $32$ commits are $11$ \textsc{Reframe}, $9$ \textsc{Refine}, $9$ \textsc{Bridge}, and $3$ \textsc{Abstract}. \textsc{Abstract} is selected $10$ times and commits $3$ times. This is a short stream, so the four counts should not be read as a stable ranking. They do not match the fixed-operator order in Table~\ref{tab:ablation_os}, where \textsc{Refine} is strongest online and \textsc{Bridge} is closest on transfer.
\end{itemize}

\subsection{Strategic Efficacy Over Time}
\label{sec:appendix_strategic_efficacy}

\begin{figure*}[t]
\centering
\includegraphics[width=\textwidth]{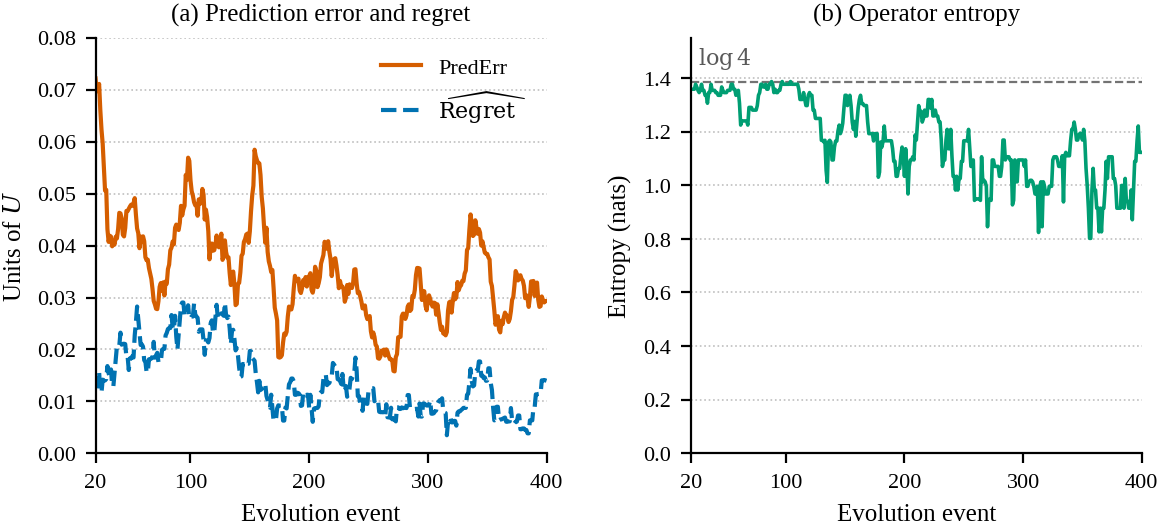}
\caption{ALFWorld strategic-efficacy terms once the window contains $K=20$ attempts (events 20--400), including \textsc{Abstract} attempts that do not commit. (a)~$\mathrm{PredErr}$ and $\widehat{\mathrm{Regret}}$ on the utility scale (Eqs.~\eqref{eq:prediction_error} and~\eqref{eq:regret_proxy}). (b)~Operator entropy $\mathrm{Div}$ in nats (Eq.~\eqref{eq:operator_diversity}). The gray line is $\log 4$, the maximum over four operators. A one-operator window has entropy $0$.}
\label{fig:strategic_efficacy}
\end{figure*}

Figure~\ref{fig:strategic_efficacy} plots the three windowed terms in Eq.~\eqref{eq:strat_eff} on ALFWorld, including \textsc{Abstract} attempts that do not commit. The window is $K=20$. Event $i$ is drawn only when $m_i=K$, so the curves start at event 20 and each point averages 20 attempts. The summary below compares events 20--29 with the last ten attempts.

On ALFWorld ($400$ attempts), $\mathrm{PredErr}$ falls from $0.063$ to $0.030$. $\widehat{\mathrm{Regret}}$ is $0.014$ on events 20--29 and $0.013$ on the last ten attempts; over events 200--400 its mean is $0.010$. $\mathrm{Div}$ starts at $1.36$, close to $\log 4 \approx 1.39$, and ends at $1.08$. The drop lines up with Figure~\ref{fig:operator_dist}a: \textsc{Abstract} stops committing, and the window moves from a four-operator mix toward \textsc{Reframe} and \textsc{Refine}. Entropy $1.08$ is still above a single-operator window, which would be $0$. The combined signal $E_i$ stays positive, moving from $0.33$ to $0.28$.

On OS there are $39$ attempts and $7$ updates, and $20$ attempts have a full window. The same comparison gives $\mathrm{PredErr}$ $0.055$ to $0.036$, $\widehat{\mathrm{Regret}}$ $0.016$ to $0.021$, and $\mathrm{Div}$ $1.36$ to $1.33$. With only twenty full-window attempts, OS is reported here as these means. $E_i$ stays near $0.34$.

The curves stop at the ALFWorld evolution cutoff. They do not cover the final $20\%$ of the online stream, where evolution is disabled. Table~\ref{tab:last20} is that later interval on the main evaluation runs. It is not a continuation of Figure~\ref{fig:strategic_efficacy}.

\subsection{Qualitative Edits}
\label{sec:appendix_qualitative}

\begin{figure*}[t]
    \centering
    \includegraphics[width=\textwidth]{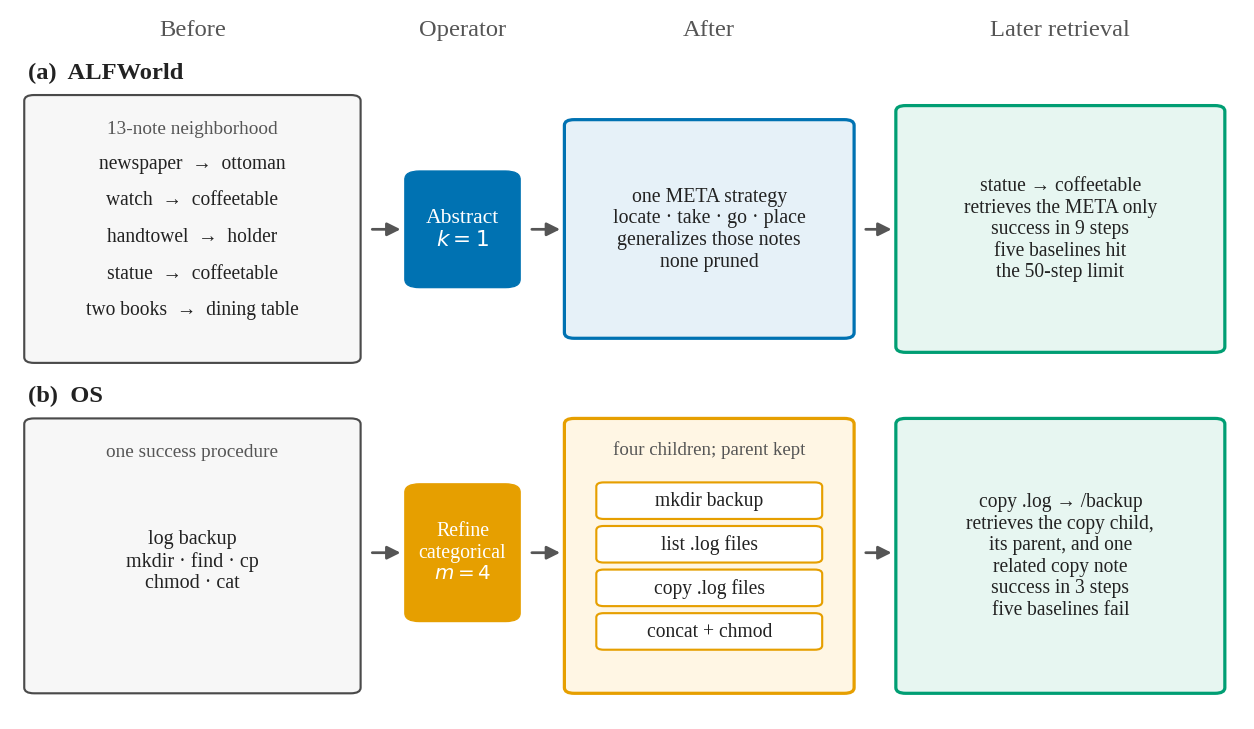}
    \caption{
        Two applied edits from the Table~\ref{tab:main} runs, read left to right:
        the graph fragment before the edit, the operator that was applied,
        the structure written into the graph, and a later retrieval.
        (a)~ALFWorld.
        \textsc{Abstract} ($k{=}1$) links a meta-strategy to a 13-note
        neighborhood and prunes nothing. Five of the six linked notes remain
        in the frozen graph and are shown.
        At stream position 1579, a new statue-to-coffeetable episode
        retrieves only that strategy and succeeds in 9 steps.
        The same game reaches the 50-step limit under No Memory, RAG,
        Self-RAG, Mem0, and MemP.
        (b)~OS.
        \textsc{Refine} (categorical, $m{=}4$) splits one log-backup
        procedure into four children and keeps the parent.
        At stream position 170, the instruction to copy \texttt{.log} files
        from \texttt{/var/log} into \texttt{/backup} retrieves the copy
        child, that parent, and one other copy note, and succeeds in 3 steps.
        The same five systems fail that instruction.
    }
    \label{fig:qualitative}
\end{figure*}

Figure~\ref{fig:qualitative} traces two applied edits from the runs in Table~\ref{tab:main}. For each edit we record the graph fragment before the change, the operator and the arguments that were applied, the nodes written into the graph, and a later episode that retrieved one of those nodes.

\textbf{ALFWorld.}
At stream position 960 the applied operator is \textsc{Abstract}, with neighborhood radius $k{=}1$ and pruning threshold $0$. The seed is a successful pick-and-place procedure for one concrete scene: go to shelf 1, take statue 1, go to coffeetable 1, and move the statue onto the table. Its 1-hop relation neighborhood contains 13 notes. The edit writes a single \texttt{META\_STRATEGY} and links it to six of those notes. No note is removed.

Five of the six linked notes remain in the frozen graph. They are separate successful procedures: a newspaper from an armchair to an ottoman, a watch to a coffeetable, a handtowel from a holder, a statue to a coffeetable, and two books to a dining table. The new strategy keeps the statue-to-coffeetable goal. Its procedure is the placement loop shared by that neighborhood. Search shelves, tables, cabinets, and drawers; open a closed source before taking; use an object index when several objects share a name; confirm that the statue is in hand; clear anything blocking the coffeetable; issue the place command only while standing at the table; then inspect the table and repeat the placement if the statue is absent. The seed procedure names one shelf, one statue index, and one table. The meta-strategy states the closed-container, index, blocker, and verification checks in the same procedure.

That meta-strategy is the only retrieved note on five later statue-to-coffeetable episodes, at stream positions 1347, 1387, 1539, 1545, and 1579. All five succeed, in 11, 8, 4, 6, and 9 steps. On the episode at position 1579, retrieval returns only that strategy. No Memory, RAG, Self-RAG, Mem0, and MemP each reach the 50-step limit on the same game without success.

\textbf{OS.}
At stream position 160 the applied operator is \textsc{Refine}, with a categorical decomposition into $m{=}4$ children. The target is one successful procedure tagged \texttt{perms}, whose command list is \texttt{find}, \texttt{sort}, \texttt{mkdir}, \texttt{cp}, \texttt{cat}, \texttt{chmod}, and \texttt{echo}. The note performs five roles in sequence. It creates \texttt{/var/log/app/backup}. It lists non-recursive \texttt{.log} files under \texttt{/var/log/app/errors} and \texttt{/var/log/app/debug} in sorted, null-separated order. It copies each file into the backup directory with \texttt{cp --preserve=mode,timestamps --backup=numbered}. It concatenates that same list into \texttt{/var/log/app/full\_report.log}. It sets the report to mode 644. The parent note is kept.

The four children split those roles and receive their own skill tags. The first, tagged \texttt{files}, only runs \texttt{mkdir -p} on the backup directory. The second, tagged \texttt{text}, only emits the sorted \texttt{find} list. The third, tagged \texttt{files}, copies the listed logs, including the numbered-backup \texttt{cp}. The fourth, tagged \texttt{perms}, concatenates the list into the report and applies \texttt{chmod 644}. Each child is tied to the parent by an \texttt{is-a} edge from the parent and a \texttt{generalizes} edge from the child.

At stream position 170 the instruction is to copy every \texttt{.log} file from \texttt{/var/log} into \texttt{/backup}, preserving directory structure, and to write a text file that lists each copied name and size. Retrieval returns three notes. The copy child has activation $0.611$, the original parent has activation $0.614$, and a separate procedure for copying \texttt{.txt} files from \texttt{/data} into \texttt{/backup} has activation $0.509$. The listing child and the concatenate-and-chmod child are absent. The episode succeeds in 3 steps. On the same instruction, No Memory fails in 4 steps, RAG in 3, Self-RAG in 5, Mem0 in 4, and MemP in 2.

The ALFWorld episode receives one placement strategy. The OS episode receives the copy step of a note that also listed files, built a report, and changed permissions. In both cases the paired baseline failures are outcomes on the same task. They do not isolate the edit from the other contents of the graph.

\clearpage
\section{Prompt Templates}
\label{sec:appendix_prompts}

Relation extraction, \textsc{Abstract}, \textsc{Bridge}, and \textsc{Reframe} use the templates below on ALFWorld, WebShop, OS, and DB.
\textsc{Refine} uses the template below on ALFWorld, WebShop, and DB.
The OS system in Table~\ref{tab:main} asks for at most \(m\) source-grounded shell subprocedures and reads them as a JSON array of \texttt{intent} and \texttt{procedure} fields.
The agent instruction and the episode-note writer are shown for ALFWorld.
WebShop, OS, and DB use the same two calls, with that benchmark's action syntax and note instructions.
Each system message fixes the role of the call and the fields the implementation reads.
Each user message supplies the current episode, the retrieved notes, or the graph fragment chosen by the executor.
Text in braces is a placeholder filled at runtime.
A note-producing call uses the five fields \texttt{TYPE}, \texttt{TASK\_TYPE}, \texttt{ROOT\_CAUSE}, \texttt{LESSON}, and \texttt{PROCEDURE}.

\subsection{Agent Step}
\label{sec:appendix_prompt_agent}

At every environment step the agent receives the task, the notes retrieved for that task, and the interaction so far, and returns one next action.
On the first two steps the user message also includes one instructional example for the current task type.
The box below is the ALFWorld instruction.
WebShop replaces the action list with \texttt{search} and \texttt{click}.
OS returns a bash block under \texttt{Act: bash} or \texttt{Act: finish}.
DB returns one SQL statement under \texttt{Action: Operation}, or a final answer under \texttt{Action: Answer}.

\begin{tcolorbox}[promptbox, title={Agent Step (ALFWorld)}]
\textbf{System.}
You solve a household task one action at a time.
The environment describes the current scene and the goal, then returns an observation after each action.
When the observation is ``Nothing happened'', the previous action was invalid.
Place an object with \texttt{move \{obj\} to \{recep\}}.
The legal actions are:
\begin{enumerate}
    \item \texttt{go to \{recep\}}
    \item \texttt{take \{obj\} from \{recep\}}
    \item \texttt{move \{obj\} to \{recep\}}
    \item \texttt{open \{recep\}}
    \item \texttt{close \{recep\}}
    \item \texttt{toggle \{obj\} \{recep\}}
    \item \texttt{clean \{obj\} with \{recep\}}
    \item \texttt{heat \{obj\} with \{recep\}}
    \item \texttt{cool \{obj\} with \{recep\}}
\end{enumerate}
Fill both fields below.
\begin{enumerate}
    \item \texttt{Thought}: the current state and the plan for the next action.
    \item \texttt{Action}: one legal action.
\end{enumerate}
\textbf{User.}\\
\texttt{Primary Goal:}\\
\{task\}\\
\texttt{Instructional Example:}\\
\{example\}\\
\texttt{Archived Memories:}\\
\{memories\}\\
\texttt{Current Task Progress:}\\
\{history\}\\
Now, it's your turn.
\end{tcolorbox}

\subsection{Episode Note}
\label{sec:appendix_prompt_note}

After an episode, the writer turns the task, the outcome, and the trajectory into one note, which is stored as a node.
The box below is the ALFWorld writer.
WebShop, OS, and DB keep these five fields and this user message.
Their system messages name shopping, shell, or SQL categories, and ask for \texttt{search}/\texttt{click} steps, concrete shell commands, or a SQL query shape.

\begin{tcolorbox}[promptbox, title={Episode Note (ALFWorld)}]
\textbf{System.}
You write one reusable memory note from a completed household episode.
Fill every field below.
\begin{enumerate}
    \item \texttt{TYPE}: \texttt{SUCCESS\_PROCEDURE}, \texttt{FAILURE\_REFLECTION}, or \texttt{META\_STRATEGY}.
    \item \texttt{TASK\_TYPE}: a household category, such as \texttt{pick\_and\_place}, \texttt{heat}, or \texttt{clean}.
    \item \texttt{ROOT\_CAUSE}: why the episode failed, or \texttt{N/A} when it succeeded.
    \item \texttt{LESSON}: one lesson that could apply to a later episode.
    \item \texttt{PROCEDURE}: numbered high-level steps. Keep them environment-agnostic where possible.
\end{enumerate}
\textbf{User.}\\
\texttt{Task:} \{task\}\\
\texttt{Outcome:} \{SUCCESS or FAILURE\}\\
\texttt{Trajectory:}\\
\{trajectory\}\\
Write a compact reusable memory note.
\end{tcolorbox}

\subsection{Relation Extraction}
\label{sec:appendix_prompt_relations}

When a new note is written, a second call proposes typed edges to notes already in the graph.
The system message fixes the relation vocabulary and the block format.
The user message is the new note and a short list of candidate notes.
An empty answer is a valid result when no relation is supported.

\begin{tcolorbox}[promptbox, title={Relation Extraction}]
\textbf{System.}
You propose typed edges between a new memory note and existing candidate notes.
Prefer \texttt{causes}, \texttt{precedes}, or \texttt{is-a}.
Use \texttt{analogous-to} only when two notes are variants of the same procedure, and then set the weight at most $0.4$.
Emit at most two blocks.
Leave the answer empty when every candidate link would be a weak guess.
Each block uses the fields below. Use only identifiers that appear in the user message, and connect \texttt{NEW} to a candidate.
\begin{enumerate}
    \item \texttt{REL}: \texttt{causes}, \texttt{precedes}, \texttt{is-a}, or \texttt{analogous-to}.
    \item \texttt{SRC}: \texttt{NEW} or a candidate identifier.
    \item \texttt{DST}: \texttt{NEW} or a candidate identifier.
    \item \texttt{W}: a weight in $[0.3, 1.0]$.
\end{enumerate}
\textbf{User.}\\
\texttt{NEW:}\\
\{new\_note\}\\
\texttt{CANDIDATES:}\\
\{candidates\}\\
Extract relations.
\end{tcolorbox}

\subsection{Abstract}
\label{sec:appendix_prompt_abstract}

\textsc{Abstract} receives the seed neighborhood selected by the policy and writes one higher-level concept.
The new node is stored as a \texttt{META\_STRATEGY} and linked to the neighborhood.
The system message asks for a single shared pattern; the user message is the seed hint and the neighborhood notes.

\begin{tcolorbox}[promptbox, title={Abstract}]
\textbf{System.}
You compress several related memory notes into one higher-level concept.
Fill every field below.
\begin{enumerate}
    \item \texttt{TYPE}: \texttt{META\_STRATEGY}.
    \item \texttt{TASK\_TYPE}: a short name for the shared pattern.
    \item \texttt{ROOT\_CAUSE}: \texttt{N/A}.
    \item \texttt{LESSON}: the theme common to the neighborhood. Drop episode-specific names unless they are hard constraints.
    \item \texttt{PROCEDURE}: the procedure shared by those notes.
\end{enumerate}
\textbf{User.}\\
\texttt{Seed:} \{seed\}\\
\texttt{Neighborhood notes:}\\
\{notes\}\\
Write one abstractive concept node.
\end{tcolorbox}

\subsection{Refine}
\label{sec:appendix_prompt_refine}

\textsc{Refine} splits one overloaded note into \(m\) children along the sampled dimension, which is temporal, causal, or categorical.
Each child is a separate note.
The system message fixes the split; the user message is the source note together with \(m\) and the dimension.

\begin{tcolorbox}[promptbox, title={Refine}]
\textbf{System.}
You decompose one overloaded memory note into finer-grained child notes.
Return exactly the requested number of children.
Split along the requested dimension: temporal, causal, or categorical.
Keep the children complementary.
Each child uses the same five fields as a memory note, under the header \texttt{CHILD\_k}.
\begin{enumerate}
    \item \texttt{CHILD\_1}: \texttt{TYPE}, \texttt{TASK\_TYPE}, \texttt{ROOT\_CAUSE}, \texttt{LESSON}, \texttt{PROCEDURE}.
    \item \texttt{CHILD\_2}: the same fields.
    \item Continue through \texttt{CHILD\_m}.
\end{enumerate}
\textbf{User.}\\
\texttt{Decomposition dimension:} \{dimension\}\\
\texttt{Number of children:} \{m\}\\
\texttt{Overloaded node:}\\
\{note\}\\
Emit \texttt{CHILD\_1} through \texttt{CHILD\_m}.
\end{tcolorbox}

\subsection{Bridge}
\label{sec:appendix_prompt_bridge}

\textsc{Bridge} receives two weakly connected regions and returns one relation type together with a short note that states how the two regions are used together.
The system message restricts the relation to the fixed vocabulary; the user message is the two clusters.

\begin{tcolorbox}[promptbox, title={Bridge}]
\textbf{System.}
You connect two memory regions that are semantically related but weakly linked.
Prefer \texttt{causes}, \texttt{precedes}, or \texttt{is-a}.
Use \texttt{analogous-to} when the two regions are variants of the same procedure.
Fill every field below.
\begin{enumerate}
    \item \texttt{RELATION}: \texttt{causes}, \texttt{precedes}, \texttt{is-a}, or \texttt{analogous-to}.
    \item \texttt{JUSTIFICATION}: one sentence.
    \item \texttt{BRIDGE\_NOTE}: a note with \texttt{TYPE} \texttt{META\_STRATEGY}, a short \texttt{TASK\_TYPE}, \texttt{ROOT\_CAUSE} \texttt{N/A}, a \texttt{LESSON} stating the link between the two regions, and a \texttt{PROCEDURE} stating how to reuse them together.
\end{enumerate}
\textbf{User.}\\
\texttt{Cluster A:}\\
\{cluster\_a\}\\
\texttt{Cluster B:}\\
\{cluster\_b\}\\
Infer the missing relation and write a short bridging note.
\end{tcolorbox}

\subsection{Reframe}
\label{sec:appendix_prompt_reframe}

\textsc{Reframe} receives an existing path and the logical frame sampled by the policy.
The user message is the path, in its current order, and that frame.
The system message asks for \texttt{FRAME}, \texttt{REORIENT}, and \texttt{SUMMARY}.
Section~\ref{sec:appendix_operators} states which of these fields the executor applies: the sampled frame overrides \texttt{FRAME}, \texttt{REORIENT} selects the edge edit, and \texttt{SUMMARY} is not written into the graph.
Node text is left unchanged.
A \texttt{reverse} reorientation, or the frame \texttt{effect\_cause}, reverses \texttt{causes} and \texttt{precedes} edges on the path.
The box below is the prompt sent to the model.
Its instruction to keep the underlying facts constrains that generation; the graph edit is the edge rule in Section~\ref{sec:appendix_operators}.

\begin{tcolorbox}[promptbox, title={Reframe}]
\textbf{System.}
You reorganize the logical frame of a reasoning chain.
Keep the underlying facts and change the retrieval-oriented order.
Fill every field below.
\begin{enumerate}
    \item \texttt{FRAME}: \texttt{temporal\_to\_causal}, \texttt{causal\_to\_temporal}, or \texttt{effect\_cause}.
    \item \texttt{REORIENT}: \texttt{reverse}, \texttt{relabel\_causes}, or \texttt{relabel\_precedes}.
    \item \texttt{SUMMARY}: a note with \texttt{TYPE} \texttt{SUCCESS\_PROCEDURE} or \texttt{META\_STRATEGY}, a short \texttt{TASK\_TYPE}, \texttt{ROOT\_CAUSE} \texttt{N/A}, a \texttt{LESSON} stating what the new frame emphasizes, and a \texttt{PROCEDURE} giving the same steps in the new order.
\end{enumerate}
\textbf{User.}\\
\texttt{Target logical frame:} \{frame\}\\
\texttt{Reasoning-chain notes, in the current order:}\\
\{notes\}\\
Reorganize the logical skeleton.
\end{tcolorbox}

\end{document}